\documentclass[twocolumn,5p]{elsarticle}

\usepackage{hyperref}
\usepackage{framed,multirow}

\usepackage{amssymb}
\usepackage{latexsym}
\usepackage{amsmath,amssymb}
\usepackage{subfigure}
\usepackage{url}
\usepackage{xcolor}
\definecolor{newcolor}{rgb}{.8,.349,.1}

\newcommand{\xt}{\textcolor[rgb]{0.00,0.00,0.0}}

\journal{Pattern Recognition}

\begin{document}

\begin{frontmatter}

%\title{Elsevier \LaTeX\ template\tnoteref{mytitlenote}}
\title{Person Re-Identification by Unsupervised Video Matching}

\author{Xiaolong Ma\fnref{Tsinghua,CAEIT}}
\ead{goup000@163.com}
\fntext[Tsinghua]{Tsinghua University, China}
%\fntext[CAEIT]{China Academy of Electronics and Information Technology}

\author[QMUL]{Xiatian Zhu}
\fntext[QMUL]{Queen Mary University of London, United Kingdom}
\ead{xiatian.zhu@qmul.ac.uk}
%\ead[url]{http://www.eecs.qmul.ac.uk/~xz303/}

\author[QMUL]{Shaogang Gong}
\ead{s.gong@qmul.ac.uk}
%\ead[url]{http://www.eecs.qmul.ac.uk/~sgg/}

%\author[Tsinghua]{Xudong Xie\corref{mycorrespondingauthor}}
%\ead{xxx@xxx}
\author[Tsinghua]{Xudong Xie}
\ead{xdxie@mail.tsinghua.edu.cn}

\author[Tsinghua]{Jianming Hu}
\ead{hujm@mail.tsinghua.edu.cn}

\author[Polytechnic]{Kin-Man Lam}
\fntext[Polytechnic]{The Hong Kong Polytechnic University, Hong Kong}
\ead{kin.man.lam@polyu.edu.hk}

\author[Tsinghua]{Yisheng Zhong}
\ead{zys-dau@mail.tsinghua.edu.cn}

\fntext[CAEIT]{China Academy of Electronics and Information Technology}



\begin{abstract}
Most existing person re-identification (ReID) methods rely only on the
spatial appearance information from either one or multiple person
images, whilst ignore the space-time cues readily available in video or image-sequence data.
Moreover, they often assume the availability of
exhaustively labelled cross-view pairwise data for every camera pair,
making them non-scalable to ReID applications in
real-world large scale camera networks.
In this work, we introduce a novel video based person ReID method
capable of accurately matching people across views from arbitrary {\em unaligned} image-sequences without any
labelled pairwise data.
Specifically, we introduce a new space-time person representation
by encoding multiple granularities of spatio-temporal dynamics
in form of time series.
Moreover, a Time Shift Dynamic Time Warping (TS-DTW) model is derived for
performing automatically alignment whilst
achieving data selection and matching between
inherently inaccurate and incomplete sequences in a unified way.
We further extend the TS-DTW model for accommodating multiple feature-sequences of an image-sequence in order to fuse information from different descriptions.
Crucially, this
model does not require pairwise labelled training data (i.e. unsupervised) therefore
readily scalable to large scale camera networks of arbitrary camera
pairs without the need for exhaustive data annotation 
for every camera pair. We
show the effectiveness and advantages of the proposed method by extensive comparisons
with related state-of-the-art approaches using two benchmarking ReID
datasets, PRID$2011$ and iLIDS-VID.
\end{abstract}


\end{frontmatter}

%\linenumbers

\section{Introduction}
\label{sec:intro}

In visual surveillance,
associating automatically individual people
across disjoint camera views
is essential.
This task is known as \textit{person re-identification} (ReID).
Cross-view person ReID enables automated discovery and analysis of person-specific long-term
structural activities over widely expanded areas and
is fundamental to many important surveillance applications
such as multi-camera people tracking and forensic search.
Specifically, for performing person ReID,
one matches a probe (or query) person observed in one camera view
against a set of gallery people captured in another disjoint view
for generating a ranked list
according to their matching distance or similarity \cite{gong2014person}.
%
%%%%%%%%%%%%%%%%%%%%%%%%%%%%%%
This is an inherently challenging problem \cite{gong2014re}.
%
% appearance only
Most existing approaches~\cite{farenzena2010person,ProsserEtAlBMVC:10,hirzer2012relaxed,zhao2013unsupervised,bhuiyan2014person,liu2014fly,zheng2016towards,zhang2016learning}
perform ReID by modelling spatial visual appearance (shape, texture and colour)
of one or multiple person images. 
However, people appearance is intrinsically limited
due to the inevitable visual ambiguity and unreliability caused by
appearance similarity among different people and
appearance variations of the same person from unknown significant cross-view changes in
human pose, viewpoint, illumination, occlusion, and dynamic background clutter.
This motivates the need of seeking additional visual information sources for person ReID.

\begin{figure}
	\centering
	\includegraphics[width=1\linewidth]{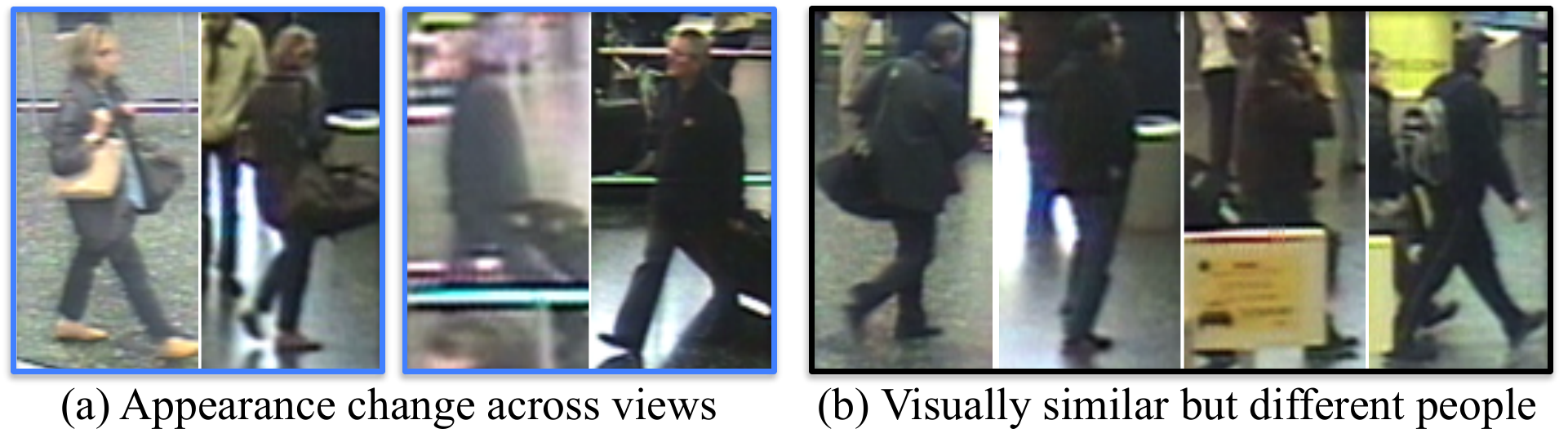}
	\vskip -.2cm
	\caption{ 
		The challenges of person re-identification in visual surveillance
		\cite{gong2014re}.
		(a) The appearance of the same person may change significantly
		across disjoint camera views
		due to great cross-camera variations in
		illumination, viewpoint,
		random inter-object occlusion and
		complex background clutter
		in typically-crowded public spaces.
		Each blue bounding box corresponds to a specific person.
		(b) Different people may present largely similar visual appearance.
	}
	\label{fig:challenges}
	% \vspace{-.3cm}
\end{figure}

%%%%%%%%%%%%%%%%%%%%%%%%%%%%%%%
% space-time information
On the other hand, video (or image-sequence) data
are often available from visual surveillance cameras.
%
% ======= Video for action recognition ========
Videos have been extensively exploited for performing action and activity recognition
by extracting and modelling a variety of dynamic space-time visual features \cite{wang2009evaluation,poppe2010survey}.
However, action recognition differs fundamentally from person ReID.
First, it often aims to
discriminate between different action categories but
tolerate the variance of the same action performed by different people.
In contrast, the objective of ReID is to discriminate
among different person identities
regardless of actions by the person.
Moreover, action recognition methods often consider a pre-defined set of
action categories during both training/testing phases,
whereas person ReID models are required to generalise
from the training categories (identities)
to previously unseen ones.

% ======= Video for gait recognition ========
Apart from action recognition, another closely related problem is gait recognition
\cite{sarkar2005humanid}.
Similar to ReID, gait recognition aims for differentiating between distinct people
by characterising people's walking dynamics.
Further, an advantage of gait recognition is
no assumption being made on either subject cooperation or
person distinctive actions.
These characteristics are analogous in spirit to person ReID.
Nonetheless, existing
gait recognition methods are heavily subject to stringent requirements on
person foreground segmentation and accurate temporal alignment
throughout a gait image sequence (a walking cycle).
%It is often assumed that complete
%gait/walking periods are captured in the target image-sequences~\cite{Han2006GEI}. 
Additionally, most gait
recognition methods do not deal
well with cluttered background and/or random occlusions with
unknown covariate conditions \cite{BashirEtAl:PRL10}
(Figures~\ref{fig:challenges} and \ref{fig:GEI}).
Hence,
person ReID in public spaces
is inherently challenging for existing gait
recognition techniques.

%% ======= Videos for ReID ========
%More recently, Wang et al. \cite{WangDVRpami} showed great benefits of
%video space-time information in assisting person ReID.
%%as well as the significant complementary effect to
%%many conventional appearance based ReID models.
%%
%However, their fragment-based Discriminative Video Ranking (DVR)
%model is limited as only a few local fragments from each
%image-sequence is exploited whilst the remaining data is totally discarded.
%Critically, the DVR model is supervised, i.e. its model training requires
%a large number of cross-view matched people for each camera pair.
%%
%This renders DVR non-scalable for
%large-scale networks with many camera pairs.

%% ======= aim of this work, challenges =======
This work aims to develop a video based person ReID approach,
without the need for exhaustively labelling people pairs across camera views.
To that end, one needs to extract and model reliably
person-specific space-time information from videos.
This is non-trivial,
especially when the videos are captured from uncontrolled and crowded public scenes.
The specific challenges include:
(1) The starting/ending frames of individual videos may correspond to
arbitrary walking phases.
Thus, any two compared videos are mostly unaligned.
This misalignment leads to inaccuracy in people matching,
especially when the useful space-time information in person videos can be very subtle.
(2) Person videos have varying numbers of walking cycles
and a holistic matching between videos may yield suboptimal recognition.
While pose estimation and walking cycle detection may help in theory,
contemporary techniques
\cite{yang2013articulated,ouyang2014multi} are still rather unreliable
for video data with distracting background and low imaging quality.
(3) Person image-sequences captured from public places
can consist of corrupted frames
due to background clutter and random inter-object
occlusions (see Figure~\ref{fig:challenges}).
A blind trust and utilisation of all visual data may
degrade the person matching accuracy.
Following \cite{wang2014person}, we call
this \textit{unregulated} image-sequences.
We wish to develop an accurate person ReID method
that does not require performing
explicit walking phase detection for videos
neither occlusion estimation for image frames.
%
%
%%%%%%%%%%%%%%%%%%%%%%%%%%%%%%
% contributions
The \textbf{main contributions} of this study are:
\begin{enumerate}
	\vspace{-.2cm}
	\item We propose an unsupervised approach to person ReID based on
	typical surveillance image-sequences.
	Our model differs significantly from
	most conventional static image based methods
	(e.g. leveraging dynamic space-time information
	{\em versus} static appearance information),
	and also the recent DVR video ReID model \cite{WangDVRpami}
	(e.g. unsupervised {\em versus} supervised).
	\item We present a new video representation particularly tailored
	for person ReID.
	Specifically, this representation is
	built up on existing action space-time features
	(e.g. histograms of oriented 3D spatio-temporal gradient \cite{klaser2008spatio})
	% histograms of optic flow \cite{laptev2008learning},
	% motion boundary histograms \cite{wang2015dense},
	and spatio-temporal pyramids \cite{lazebnik2006beyond,pirsiavash2012detecting}.
	In contrast to most visual features for action recognition which are vectorial,
	our video representation is in form of sequence or time series.
	This is specially designed for reliable selection based person matching
	between cross-view unregulated video pairs
	with possibly ambiguous, incomplete and noisy observation.
	\item We introduce an effective video matching algorithm,
	Time Shift Dynamic Time Warping (TS-DTW)
	and its Multi-Dimensional variant MDTS-DTW,
	for data selective based sequence matching.
	Particularly,
	the proposed model computes the distance between two videos
	by iteratively (1) altering their mutual time shift relation and (2)
	then matching two partial segments of them.
	Importantly, our method is capable of simultaneously
	performing sequence alignment, selecting best-matched segments,
	and fusing diverse information
	for person ReID in a unified manner.
\end{enumerate}

We show the effectiveness of the proposed approach
on two benchmarking image-sequence ReID datasets
(PRID$2011$ \cite{hirzer11a} and iLIDS-VID \cite{wang2014person})
under both the closed-world and
more realistic open-world scenarios
\cite{liao2014open,zheng2016towards}.
Extensive comparative evaluations were conducted
by comparing alternative sequence-matching person recognition models
including gait recognition \cite{martin2012gait} and
dynamic time warping \cite{rabiner1993fundamentals},
and the state-of-the-art person ReID methods including
SDALF \cite{farenzena2010person},
eSDC \cite{zhao2013unsupervised},
DVR \cite{WangDVRpami},
RDL \cite{ElyorBMVC15},
and XQDA \cite{liao2015person}.

\xt{The remainder of this paper is organised as follows.
In Section \ref{sec:related_work},
we discuss broadly the related studies.
In Section \ref{sec:overview},
we present an overview of our approach,
followed by video representation
in Section \ref{sec:representation},
video matching in Section \ref{sec:model},
and person re-identification application
in Section \ref{sec:ReID}.
Then, we depict the experimental settings
in Section \ref{sec:Exp_Settings} and provide
comparative evaluations of our proposed approach
in Section \ref{sec:Exp}.
Finally, we conclude this study %and indicate future research directions 
in Section \ref{sec:conclusion}.}

%------------------------------------------------------------------------------------------------
\section{Related Work}
\label{sec:related_work}

\begin{figure} %[h] %[!htbp]
	\centering
	%\vspace{-0.3cm}
	\subfigure %[PRID2011]
	{
		\includegraphics[width=0.99\linewidth]{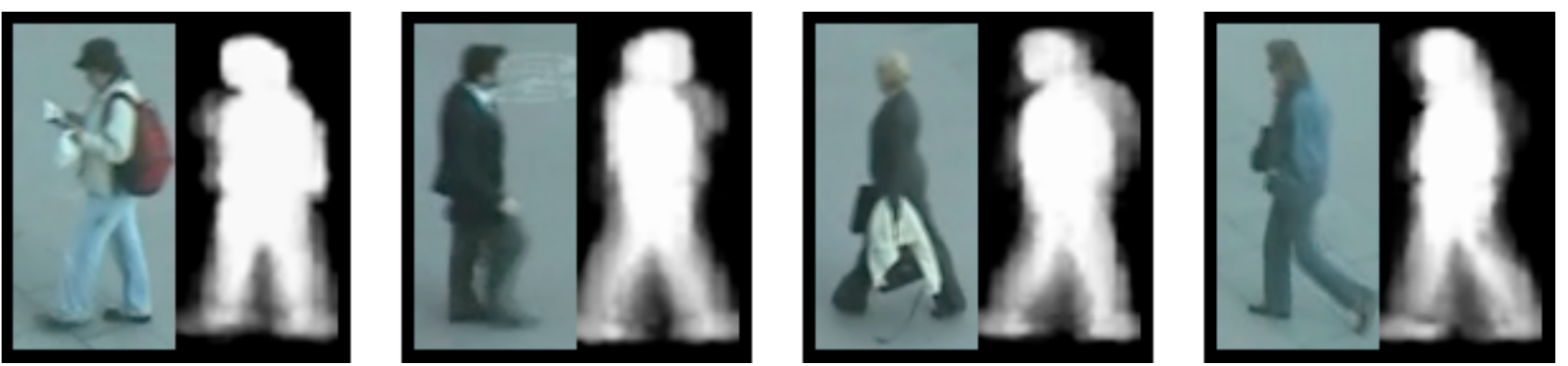}%gei4.pdf
	}
	\vskip -.2cm
	\subfigure %[iLIDS-VID]
	{
		\includegraphics[width=0.99\linewidth]{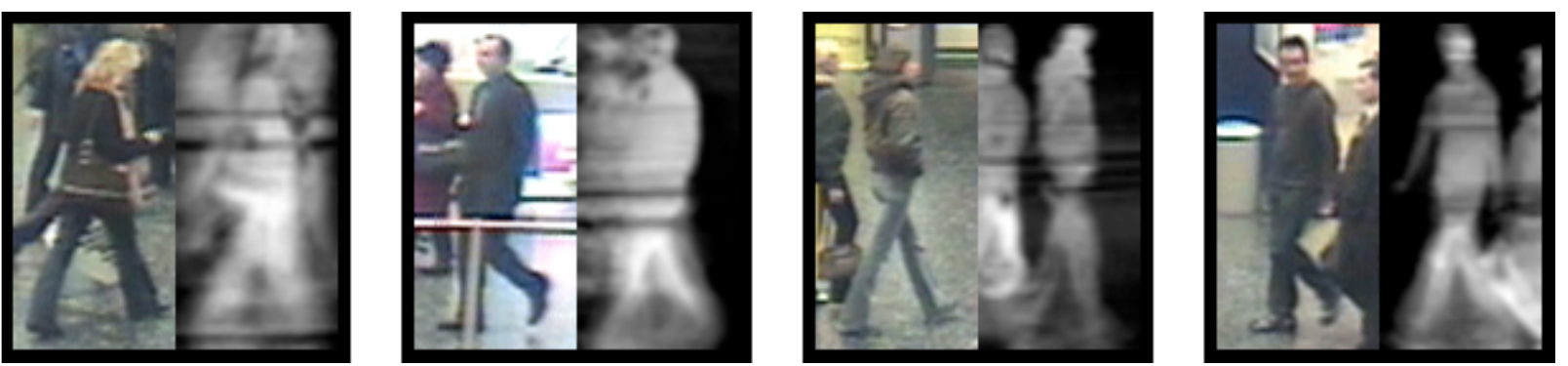}%gei4.pdf
	}
	\vskip -.2cm
	\caption{
		Example GEI features of
		PRID$2011$ \cite{hirzer11a} (top) %($1^\text{st}$ row)
		and
		iLIDS-VID \cite{wang2014person} (bottom) %($2^\text{nd}$ row)
		videos.
	}
	\label{fig:GEI}
	%\vspace{-0.3cm}
\end{figure}

%--------------------------------------------
\noindent {\bf Gait recognition.}
Gait recognition
\cite{sarkar2005humanid,xu2012human,hofmann2014tum,chattopadhyay2014pose,choudhury2015robust}
has been extensively exploited for people identification
using video space-time features, e.g.
correlation based motion feature \cite{Otsu_ICPR2004},
and Gait Energy Image (GEI) templates \cite{Han2006GEI}.
To improve gait representations, Veres et al. \cite{veres2004image}
and Matovski et al. \cite{MatovskiNMM12}
suggest feature selection and quality measure.
These methods assume that image-sequences are aligned and
captured in controlled environments with uncluttered background,
as well as having complete gait cycles, little occlusion,
and accurate gait phase estimation.
However, these constraints are often invalid in person ReID context as shown in
Figures  \ref{fig:GEI} and \ref{fig:datasets}.

% occlusion
To handle often-occurring occlusion,
Hofmann et al. \cite{Hofmann_ICCGVCV2011} propose a specific dataset
for evaluating their negative influence on gait recognition performance.
%especially the dynamic one.
Meanwhile, a number of part-based methods
\cite{boulgouris2007human,hossain2010clothing,shaikh2014gait} are developed
by assuming that matched people share common observed parts (COPs).
For relaxing this assumption, Muramatsu et al. \cite{Muramatsu_ICB2015}
reconstruct complete gait features from partially
observed body parts without sharing COPs.
These methods rely on accurate body part segmentation and occlusion detection,
which is however over-demanding for contemporary segmentation methods \cite{yang2013articulated,ouyang2014multi,xiao2006bilateral}
given typical ReID video data captured against
uncooperative people and dynamic scenes.

% deal with covariates
Main challenges for gait recognition arise from
various covariate conditions,
e.g. carrying, clothing, walking surface, footwear, and viewpoint.
Beyond the attempts of designing and investigating gait features
invariable to specific covariates
\cite{sarkar2005humanid,yu2006modelling,yang2008gait,singh2009biometric,BashirEtAl:PRL10},
more powerful learning based methods have also been presented
for explicitly and accurately modelling 
the complex variances of gait structures.
For example, Mart{\'\i}n-F{\'e}lez and Xiang \cite{martin2014uncooperative}
exploit the learning-to-rank strategy for jointly
characterising a variety of covariate conditions in a unified model.
Whilst a learning process may help improve the gait recognition accuracy,
this strategy is heavily affected by the goodness of gait features.
On person ReID videos however, gait features are likely to be extremely unreliable,
as demonstrated in Figure \ref{fig:GEI}.

%------------------------------------------------
\vspace{0.1cm}
\noindent {\bf Temporal sequence matching.}
Temporal sequence matching is another alternative strategy.
The Dynamic Time Warping (DTW) model
\cite{rabiner1993fundamentals,senin2008dynamic,rakthanmanon2012searching}
and its variants including
derivative DTW \cite{keogh2001derivative,gullo2009time},
weighted DTW \cite{jeong2011weighted},
are common sequence matching algorithms widely used in
data mining and pattern recognition.
Given two temporal sequences,
it searches for the optimal non-linear warp path between the sequences
that minimises the matching distance.
However, the conventional DTW models assume that the two sequences
have the same number of temporal cycles (phases) and
are aligned at the starting and ending points/elements.
These conditions are difficult to be met in %automatically captured
person videos from typical surveillance scenes.
Hence, directly using DTW variants to holistically match
these unregulated videos may be suboptimal.
To further compound the problem, there are often unknown occlusions
and background clutters that can lead to corrupted video frames
with missing and/or noisy observation thus potentially inaccurate distance measurement.

In case of cyclic sequences, e.g. closed curves,
the starting element is often unknown and
may be located by a greedy search or some heuristic method \cite{Horng2002}.
However, there can exist more than one starting elements for periodic
sequences like people walking videos.
Whilst continuous dynamic programming or spotting \cite{oka1998spotting}
identifies both starting/ending elements,
it requires a good pre-defined threshold, which however is not available
in our person ReID problem.

%--------------------------------------------------
\vspace{0.1cm}
\noindent {\bf Single/multi-shot and video based person ReID.}
Most existing ReID methods \cite{ProsserEtAlBMVC:10,hirzer2012relaxed,zhao2013unsupervised,bhuiyan2014person,liu2014fly,wuz,chen2015mirror,wang2016human,wangtowards,kodirov2016unsupervised}
only consider one-shot image per person per view. This is inherently
weak when multi-shot %or image sequences
are available, due to
the intrinsically ambiguous and noisy people appearance and large
cross-view appearance variations (Figure \ref{fig:challenges}).
There are efforts on multi-shot ReID. For example,
Hamdoun et al. \cite{hamdoun2008person} propose to employ the interest points
cumulated across a number of images;
Cong et al. \cite{cong2009video} utilise the data manifold geometric
structures of multiple images %in image sequences
for constructing more compact spatial appearance description.
Other attempts include training a robust appearance model using image
sets~\cite{nakajima2003full} and
enhancing local image region/patch spatial feature representation~\cite{gheissari2006person,farenzena2010person,cheng2011custom,xu2013human}.
In contrast to all these methods focusing on exploiting spatial
appearance information, this work explores space-time information
from available videos for person ReID.

Previous efforts of exploiting space-time dynamics for person ReID
are built on either gait recognition or action recognition.
Specifically,
gait features are exploited for enriching appearance ReID representations in \cite{roy2012hierarchical,kawai2012person,bedagkar2014gait,liu2015enhancing}.
But these methods naturally share similar limitations of gait recognition models,
e.g. severely suffering from feature noises inherent in ReID data.
Recently, Wang et al. \cite{wang2014person,WangDVRpami} partly solve this problem
by formulating a discriminative video ranking (DVR) model
using the space-time HOG3D feature \cite{klaser2008spatio}.
\xt{However, this fragment-based DVR
model is limited as only a few local fragments from each
person image-sequence is exploited whilst the remaining data is totally discarded.
Critically, the DVR model is supervised, i.e. its model construction requires
a large number of cross-view matched people for each camera pair.
This renders DVR non-scalable for
large-scale networks with many camera pairs.}
Other video based ReID methods \cite{you2016top,mclaughlinrecurrent} are also supervised
and thus subject to the similar scalability limitation as DVR.

\vspace{0.1cm}
\noindent {\bf Space-time visual features.} %Action recognition.}
Our person video representation is inspired by
existing successful action features and the DVR model \cite{WangDVRpami},
e.g. histograms of oriented 3D spatio-temporal gradient (HOG3D) \cite{klaser2008spatio}.
%
% ******* sequences or time series ****************
%
In contrast to most feature vector based action representations
\cite{schuldt2004recognizing,dollar2005behavior,laptev2008learning,kim2009canonical,scovanner20073,willems2008efficient,pirsiavash2012detecting,Zhu2015Convolutional},
we represent person videos with temporal sequences based representations.
%ours is in form of sequence or time series.
This design is capable of (1) not only
encoding the dynamic temporal structures of motion,
(2) but also selectively matching unregulated person videos
(see Section \ref{sec:model}).
While some action recognition models also regard videos as sequences of observation \cite{nowozin2007discriminative,schindler2008action,niebles2010modeling,gaidon2011actom,gaidon2011time}, their focus is coarse temporal structure modelling alone.

%
% ******* temporal pyramids ****************
%
To extract different granularities of localised temporal ordering dynamics,
we adopt the notion of temporal pyramids % \cite{pirsiavash2012detecting,zelnik2001event,irani1996efficient}
% into our video representation,
%i.e. a set of different sequence-element lengths
(see Figure \ref{fig:sequentialisation}(b)).
Instead of using temporal sub-sampling to construct a
temporal pyramid \cite{zelnik2001event,irani1996efficient},
we segment videos with different sequence-element lengths
for preserving all possible dynamic information at all levels as in
\cite{pirsiavash2012detecting,choi2008spatio}.
However, our representation is different significantly from the latter two
%\cite{pirsiavash2012detecting} and \cite{choi2008spatio}
because:
(1) They use vector based representations {\em whilst}
ours are sequential or temporal series;
(2) They assume well segmented videos as input (e.g. one action per video) {\em whilst}
our person videos can contain a varying numbers of walking action periods
%(e.g. consider each walking cycle as an action)
without any temporal segmentation;
%
% ******* space-time pyramids ****************
%
(3) We {\em additionally} consider spatial pyramid \cite{lazebnik2006beyond}
at each temporal granularity and importantly data selection in video matching.

%------------------------------------------------------------------------------
\section{Approach Overview}
\label{sec:overview}
% === problem ===
% Eddy here =====================================
Unlike most action recognition methods that represent each video with a feature vector \cite{wang2009evaluation}
or the image-sequence based person re-identification (ReID) approach
that describes each video with a set of independent vectors \cite{wang2014person,WangDVRpami},
we consider person videos as sequences of localised space-time dynamics for
performing ReID.
This allows to:
(1) Explicitly represent and model localised temporal motion dynamics;
(2) Flexibly achieve temporal alignment between different videos;
(3) Facilitate data driven selective matching without any supervision (see Section \ref{sec:model}).
All these capabilities are desired and helpful for reliable person ReID by
accurately characterising and exploiting
space-time dynamic information
of person's walking behaviour recorded in unregulated videos
with random inter-object occlusions,
arbitrary video duration and uncertain starting/ending phases,
and uncontrolled background clutter.

\begin{figure*}
	\centering
	\includegraphics[width=1\linewidth]{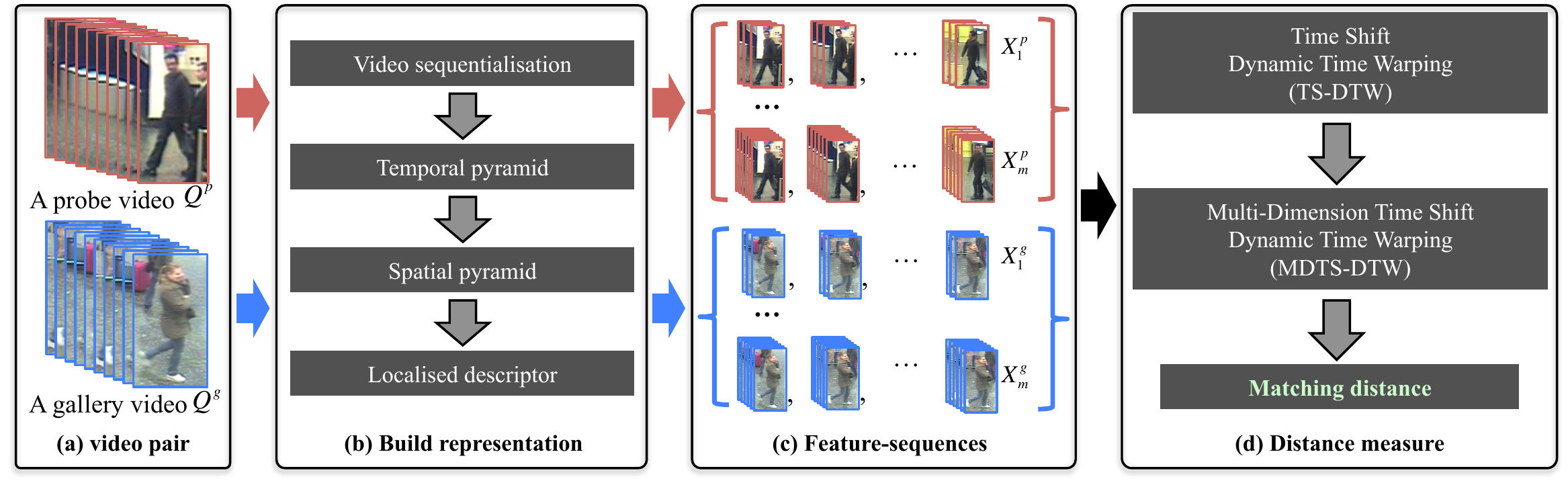}
	\vskip -.2cm
	\caption{\footnotesize
		\xt{
			Overview of the proposed unsupervised video matching approach for
			person ReID.
			(a) An input pair of person videos;
			(b) Construct video representation by
			video sequentialisation (Section \ref{sec:sequentialisation}),
			temporal pyramid (Section \ref{sec:temopral-pyramids}),
			spatial pyramid (Section \ref{sec:spatial-pyramids}), and
			localised space-time descriptor computation (Section \ref{sec:descriptor});
			(c) Obtained feature-sequences;
			(d) Video matching
			by the proposed TS-DTW (Section \ref{sec:TS-DTW}) and
			MDTS-DTW (Section \ref{sec:multi-dimension}) models.
		}
	}
	\label{fig:framework}
	%\vspace{-.3cm}
\end{figure*}

% === challenges ===
% In person ReID context,
However, it is non-trivial to automatically detect and exploit
identity-sensitive space-time information from noisy video
data, particularly in an unsupervised manner.
Critically, one needs to address the problems of
(1) how to extract rich dynamics information of people's walking motion, and
(2) how to suppress the negative influence of unknown noisy observation,
e.g. various types of occlusion and clutter in the background.
This is beyond solving the more common temporal misalignment problem in video matching.
% === our method ===
To this end, we formulate a novel unsupervised person re-identification method capable of extracting multi-scale spatio-temporal structure information (Section \ref{sec:representation}),
automatically aligning sequence pairs and
adaptively selecting/employing informative visual data (Section \ref{sec:model})
from noisy person videos captured in non-overlapping camera views.
\xt{This allows to relax the stringent assumptions
of existing gait recognition methods
and overcome the limitations of previous temporal sequence matching models,
and result in more accurate person recognition,
particularly with incomplete and noisy person videos captured in public spaces.
Compared with the state-of-the-art DVR re-id model,
our method is able to extract and employ much richer space-time
cues from videos.
Moreover, the proposed method is unsupervised,
as opposite to DVR which needs a large number of
cross-view matching pairs for every camera pair.
Therefore, our proposed method is more scalable to the real-world applications
involving large surveillance camera networks.
Additionally, we further consider information fusion
from multiple feature-sequences each capturing some different aspects of person video data.}
An overview diagram of the proposed approach is presented in Figure~\ref{fig:framework}.

%------------------------------------------------------------------------------
\section{Structured Video Representation}
\label{sec:representation}

%sssssssssssssssssssssssssssssssssssss
\subsection{Video Sequentialisation}
\label{sec:sequentialisation}

Suppose we have a collection of video (or image-sequence) pairs
$\{(Q_i^p, Q_i^g)\}_{i=1}^n$,
where $Q_i^p$ and $Q_i^g$ denote \xt{the videos of person $i$} captured
by two disjoint cameras $p$ and $g$,
and $n$ the number of people.
Each video is defined
as a set of consecutive frames $\mathbf{I}$ (e.g. obtained by
an independent person tracking process \cite{smeulders2014visual}
with simple post-processing or not):
$Q = \{\mathbf{I}_1, \mathbf{I}_2, ...\}$,
where the video length $|Q|$ is varying as in
typical surveillance settings, independently extracted person videos do not
guarantee to have a uniform duration (arbitrary frame number), nor
the number of walking cycles and starting/ending phases.

Given varying-long videos with unknown and random noise,
it is ineffective to perform matching between two image-sequences holistically.
A possible strategy \cite{WangDVRpami} is:
(1) Segmenting each video into multiple independent fragments;
(2) Selecting the optimal/best fragment pairs for matching.
This method, however, may lose potentially useful information
encoded in the discarded fragments.
In this work, we instead consider
a richer representation for exploiting as much space-time information
from inherently noisy videos as possible.

\begin{figure}
	\centering
	\includegraphics[width=0.99\linewidth]{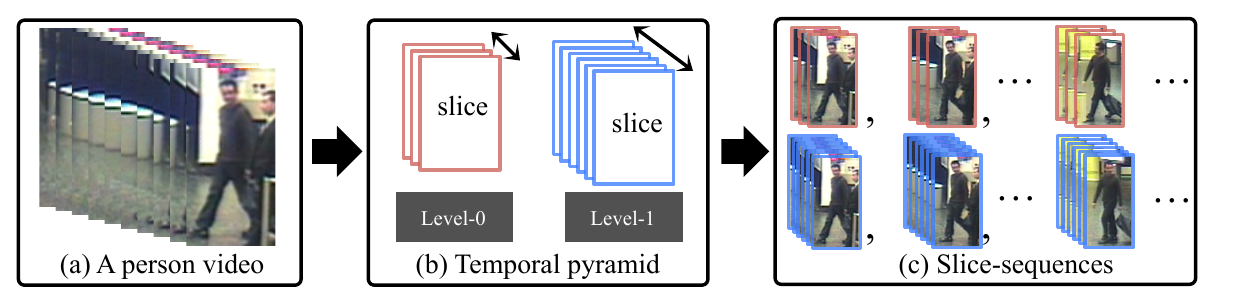}
	\vskip -0.2cm
	\caption{
		\xt{Illustration of temporal pyramid and video sequentialisation.
			Note the colour-coded correspondence between
			(b) the temporal pyramid level and (c) the slice-sequence.
		}
	}
	%\vspace{-.3cm}
	\label{fig:sequentialisation}
\end{figure}

Specifically, we divide uniformly each individual video $Q$ into
multiple temporally localised \textit{slices} with a small number $l$ of image frames.
Different slice lengths $l$ correspond to different temporal granularities.
Each slice encodes localised space-time information
about the walking characteristics of the corresponding person.
As a result, a video can be converted into
a \textit{space-time slice-sequence}
$S = \{s_1, s_2, \dots\}$ (Figure \ref{fig:sequentialisation}).
This localised slice-based sequence representation has three
advantages over the bag-of-fragments model \cite{wang2014person}:
% allows us to effectively perform
(1) It keeps the original sequential data form, whilst DVR only
considers each fragment of a sequence as an isolated instance without temporal
ordering among fragments.
This allows us to enjoy the merits of existing sequence matching algorithms,
e.g. non-linear dynamic time warping for handling the misalignment problem.
(2) Alignment between sequences (e.g. starting/ending with the same walking
phases) is made more robust due to the existence of a large
number of short localised slices corresponding to various walking
phases. In contrast, the bag-of-fragments strategy may suffer from
fragilely aligned fragment pairs at times when only a small number of
fragments are available from a video and the starting/ending phases
of fragments are not sufficiently diverse to match.
(3)
It provides more flexible opportunities for selecting and
exploring informative localised space-time information irregularly distributed
across the original image-sequences, e.g. not only in the form of
isolated fragments.
% e.g. either the whole sequence or partial segments at any locations with arbitrary lengths.
%
This is difficult for the bag-of-fragments representation in DVR due to its
hard video fragmentation and coarse fragment selection mechanism.

%sssssssssssssssssssssssssssssssssssss
\subsection{Temporal Pyramid}
\label{sec:temopral-pyramids}

Since variations in walking styles may exist over
various local temporal extends,
it is suboptimal to utilise video slices of a uniform length.
Also, fine-to-coarse localised temporal information is possible
to complement each other in expressing temporal structure dynamics, as demonstrated in
existing action recognition studies \cite{pirsiavash2012detecting,choi2008spatio}.
In light of these considerations,
we enrich our representation of person videos by imposing a temporal pyramid structure,
motivated by pyramid match kernel \cite{grauman2007pyramid}
and its spatial extension \cite{lazebnik2006beyond}.

Specifically, %instead of one single slice duration,
we use a set of video slice length for video sequentialisation as:
\begin{equation}
	L = \{2^0l, \; \dots, \; 2^{(h_t-1)}l\}
\end{equation}
which corresponds to a temporal pyramid with $h_t$ levels/layers.
Given a video $Q_i$, we generate a separate slice-sequence
at each temporal pyramid level.
Thus, a total of $h_t$ slice-sequences $\{S_i^l\}_{l=0}^{h_t-1}$ can be produced for
each video $Q_i$ after applying this temporal pyramid
(Figure \ref{fig:sequentialisation}(c)).
During sequentialising a video, at any temporal pyramid level,
we discard the last few image frames of person videos
if they are not sufficient to form a slice.
\xt{For example, suppose there is $56$ frames in a person video
	and the slice length is $10$, we drop/ignore the last $6$ frames
	as they are not enough for a complete slice of $10$ frames.}

%sssssssssssssssssssssssssssssssssssss
\subsection{Spatial Pyramid}
\label{sec:spatial-pyramids}

\begin{figure} %[h]
	\centering
	\includegraphics[width=0.6\linewidth]{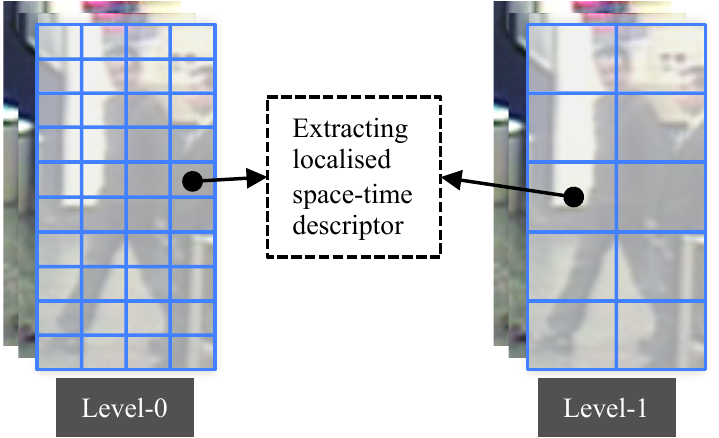}
	\vskip -.2cm
	\caption{
		Spatial pyramid structures on a temporally-localised video slice.
	}
	%\vspace{-.3cm}
	\label{fig:spatial-pyramid}
\end{figure}

After obtaining slice-sequences
$S = \{s_1, \dots, s_i, \dots\}$ of person video,
we need to consider how to represent their localised video slices $s_i$.
This is the same as
deriving video representation for action recognition \cite{wang2009evaluation}
in that each slice can be considered as a tiny action video.
We want to capture localised spatio-temporal dynamic structures of people's walking.
Apparently, the style or characteristics of walking motion is
closely related to the action of different body parts,
e.g. head, torso, arms, legs.
Hence, we spatially decompose every slice into a grid of $2$$\times$$5$ uniform cells which
approximately correspond to the layout of all body parts (Figure \ref{fig:spatial-pyramid}(right)).
This division allows to encode roughly detailed spatial cues of individual parts
into video slices.

Additionally, accurate ReID may need more fine-grained and subtle
spatially structured cues of people's walking behaviour.
This is because finer spatial decomposition provides more detailed information
and potentially complements coarse divisions.
%although at increasing risk of misalignment between corresponding regions in slices.
To that end, we adopt the spatial pyramid match kernel \cite{lazebnik2006beyond},
due to its superior expressive capability shown in action recognition \cite{laptev2008learning}.
In particular, we further split each cell into $2$$\times$$2$ smaller ones,
resulting in a grid of $40$ cells on each slice (Figure \ref{fig:spatial-pyramid}(left)).
By repeating this process, we can obtain a $h_s$-level spatial pyramid.
Together with temporal pyramid, we call our video representation as
``Spatio-Temporal Pyramidal Sequence'' ({\bf STPS}).
%
%
%After applying the spatial pyramid on slices,
Next, we describe the dynamic feature descriptor for numerically representing
localised space-time cells below.

%sssssssssssssssssssssssssssssssssssss
\subsection{Localised Space-Time Descriptor}
\label{sec:descriptor}

We consider the HOG3D feature~\cite{klaser2008spatio} 
for representing video slices
due to its strong expressiveness for recognising different
activities~\cite{shi2013lpm}
and importantly for distinguishing between distinct people~\cite{wang2014person,WangDVRpami}.
\xt{Particularly,
given a specific spatial division on any video slice $s$,
we first extract the space-time gradient histogram from each cell
where 3D gradient orientations are quantised using regular polyhedrons~\cite{klaser2008spatio},
then concatenate them to form a HOG3D feature vector $\mathbf{x}$ for the slice $s$.}
Note that there is $50\%$ overlap between any two adjacent cells
for increasing robustness against tracking/annotation errors.
As such, we obtain a HOG3D feature-sequence
$X = \{\mathbf{x}_1, \mathbf{x}_2, \dots\}$
for a slice-sequence
$S = \{s_1, s_2, \dots\}$.
Finally, we apply histogram equalisation for
reducing the effect of uneven illuminations.
While other space-time descriptors, such as
motion boundary histograms (MBH) \cite{wang2015dense}, are considerable,
it is beyond our scope to exhaustively discuss and evaluate 
a variety of different space-time descriptors.

%--------------------------------------------------------------------
\section{Unsupervised Video Matching}
\label{sec:model}

In this section, we describe the details of
the proposed sequence/video matching model
for person ReID.
We aim to formulate an unsupervised model.
As a result, the expensive cross-camera pairwise labelling process for
every camera pair can be eliminated for realising good deployment scalability in reality.
To that end, we select
the well-known Dynamic Time Warping (DTW) algorithm
\cite{rabiner1993fundamentals,berndt1994using} as the basis of our model due to:
(1) Its great success and popularity in sequence based data analysis;
(2) Its simple but elegant modelling.

Specifically, we derive a new sequence matching algorithm based on the DTW model,
called {\em Time Shift Dynamic Time Warping} (TS-DTW),
and further generalise TS-DTW to the multi-dimensional setting,
i.e. with multiple feature-sequences per person video.
This formulation is motivated by works in
time delay based studies \cite{fraser1986independent,LoyIJCV10},
multi-dimension fusion \cite{shokoohi2015non}, and
neural networks (or deep learning) \cite{krizhevsky2012imagenet}.
This proposed model is characterised with
alignment free, data selection, and information fusion.
Before detailing our method,
let us first briefly describe the conventional DTW model.

\subsection{Conventional DTW}
\label{sec:DTW}
In general, \xt{the DTW model \cite{rabiner1993fundamentals,senin2008dynamic,rakthanmanon2012searching,berndt1994using}} aims at measuring the distance or similarity between
two temporal-sequences by searching for the optimal non-linear warp path.
Formally,
given two feature-sequences
$X^p = \{s_1^p,\dots, s_i^p, \dots\}$ and
$X^g = \{s_1^g, \dots, s_j^g, \dots\}$,
we define a warp path as:
\begin{equation}
	W = \{\mathbf{w}_1, \dots, \mathbf{w}_d\}
	\label{eqn:warp_path}
\end{equation}
where
the $k$-th entry $\mathbf{w}_k = (w_k^p, w_k^g)$ indicates
that the $w_k^p$-th element from $X^p$
and $w_k^g$-th element from $X^g$ are matched.
The warp path length holds as:
$\text{max}(|X^p|, |X^g|) \leq d < |X^p| + |X^g|$.
The symbol $|\cdot|$ denotes the set size.
We then define the sequence matching distance $\text{dist}_{\mathrm{dtw}}(X^p, X^g)$ between
$X^p$ and $X^g$ as:
\begin{equation}
	\text{dist}_{\mathrm{dtw}} (X^p, X^g) = \frac{1}{d}{\sum_{k=1}^{d} \text{dist}_{\text{el}}(\mathbf{x}_{w_k^p}^p, \mathbf{x}_{w_k^g}^g)}
	\label{eqn:matching_dist}
\end{equation}
with $\text{dist}_{\text{el}}(\cdot, \cdot)$
as the distance metric between two elements (or slices),
e.g. $L_1$ or $L_2$ norm,
and $d = |W|$ the warp path length.
The objective of DTW is to find the optimal warp path $W^*$ such that
\begin{equation}
	W^* = \text{argmin}_{W \in \Omega} \text{dist}_{\mathrm{dtw}} (X^p, X^g)
	\label{eqn:DTW_dist}
\end{equation}
where $\Omega$ is the set of all possible warp paths.
This optimisation can be realised using dynamic programming~\cite{muller2007dynamic}
subject to three constraints:
(1) bounding constraint: $\mathbf{w}_1=(1,1)$ and $\mathbf{w}_d=(|X^p|, |X^g|)$;
(2) monotonicity constraint:
$w_1^p \leq w_2^p \leq ... \leq w_d^p$ and
$w_1^g \leq w_2^g \leq ... \leq w_d^g$;
and (3) step-size constraint:
$\mathbf{w}_{k+1} - \mathbf{w}_k \in {(1,0),(0,1),(1,1)}$ for $k \in [1:d-1]$.

As indicated in the above bounding constraint,
DTW assumes that the starting and ending
data elements of the two sequences are aligned.
However, this is mostly invalid in videos available
for person ReID as aforementioned.
Moreover, DTW utilises all sequence element data for distance computation,
regardless the quality of individual elements.
This is likely to make the obtained distance sensitive to data noise
often present in typical ReID videos.

%-----------------------------------------------------------------
\begin{figure*} %[!htbp]
	\centering
	{
		\includegraphics[width=1\linewidth]{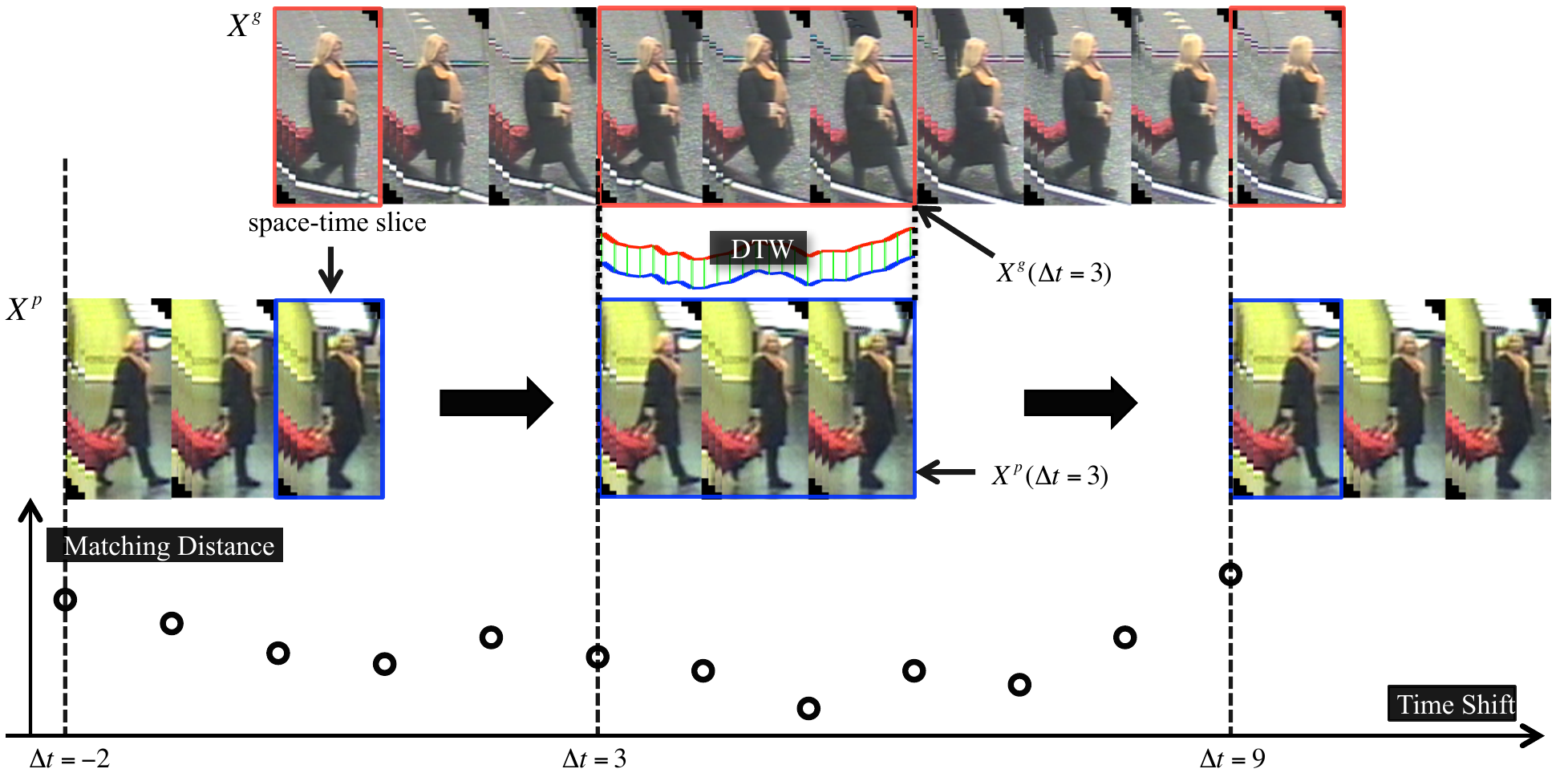}
	}
	%\vskip -1.1cm
	\caption{
		\xt{Overview of our proposed time shift driven sequence alignment and % selective
		matching.}
	}
	\label{fig:time-shift}
\end{figure*}

\subsection{Time Shift Driven Alignment and Selective Matching}
\label{sec:TS-DTW}

To overcome the above limitations of DTW,
we develop a new model,
Time Shift Dynamic Time Warping {\bf (TS-DTW)},
by introducing additionally the notions of
time shift and max-pooling into sequence matching.
Instead of matching two sequences $(X^p, X^g)$ holistically at one time as DTW,
we perform iterative and partial matching.
\xt{An illustration of this time shift driven
sequence alignment and matching is depicted
in Figure \ref{fig:time-shift}.
Specifically,
given two feature-sequences $X^p$ (probe) and $X^g$ (gallery),
we temporally shift one sequence (say $X^p$) against the other ($X^g$) from the beginning position
(where only the rightmost slice of $X^p$ is utilised
in matching with the leftmost slice of $X^g$,
e.g. $\Delta t = -2$ as in Figure \ref{fig:time-shift}),
to the ending position 
(where the rightmost slice
of $X^g$ is matched with the leftmost slice of $X^p$,
e.g. $\Delta t = 9$ as
in Figure \ref{fig:time-shift},
and black dotted vertical lines indicate several (not all) shift positions
attempted during the entire shifting process).
At any shift $\Delta t$,
the alignment between partial segments $X^p(\Delta t)$ and $X^g(\Delta t)$
(highlighted by the corresponding blue and red bounding box in Figure \ref{fig:time-shift}) is
performed by the conventional DTW algorithm \cite{berndt1994using}.
As such, a set of local matching distances
$D = \{ \text{dist}_\text{dtw}(X^p, X^g, \Delta t) \}_{\Delta t \in T}$
(indicated as the black hollow circles in Figure \ref{fig:time-shift})
can be generated over all time shifts $T$.}
Finally, we obtain the person video matching distance
by taking together all local ones as
\begin{equation}
	\text{dist}_\text{ts}(X^p, X^g) = \text{min}_{\Delta t \in T} \{ \text{dist}_\text{dtw}(X^p, X^g, \Delta t) \}
	\label{eqn:dist_TSDTW}
\end{equation}
i.e. selecting the best-matched result.
This {\em time shift ensemble} model is inspired by the max-pooling layer in neural networks
which aim at summarising the responses of neighbouring groups of neurons
\cite{krizhevsky2012imagenet}.
We cope with a similar situation if sequence-element is thought of as neuron
and sequence-segment as group of neurons.
Critically, the max-pooling operation has data selection capability for
guiding the supervised learning of neurons in neural network learning.
Whereas our objective is to achieve data selective sequence matching
or recognition in an unsupervised way,
enjoying similar spirit but with a different learning strategy.

\vspace{0.1cm}
{\underline {\em Discussion}}
The data selection capability in our proposed matching algorithm above
is significant to accurately matching sequences,
especially for unregulated ReID videos
from uncontrolled camera viewing conditions.
We summarise the key points for data selection below.
%＝＝＝＝＝＝＝＝
First, we automatically select the starting/ending walking poses, in contrast to
DTW which enforces the first and last elements of compared sequences to be aligned
so potentially introduces weak or noisy alignments into distance computation.
%
%＝＝＝＝＝＝＝＝
Moreover, we attempt many different partial segments of $X^p$ and $X^g$,
and select the best-aligned parts for distance estimation,
different from DTW that uses all observed data regardless of how good the
constituent elements are.
Thus, noisy elements can be possibly suppressed in distance computation.
These two abilities are achieved by successively varying $\Delta t$,
since the element data of $X^p(\Delta t)$ and $X^g(\Delta t)$
changes over time shifts.
Apparently, the two benefits are complementary to each other
and their combination allows us to more accurately
match incomplete and noisy surveillance videos for person ReID in an unsupervised manner,
as demonstrated by our experimental evaluations in Section \ref{sec:Exp}.

\subsection{Generalisation to the Multi-Dimensional Setting}
\label{sec:multi-dimension}

The TS-DTW model presented in Section \ref{sec:TS-DTW}
assumes one feature-sequence per person video.
This is the single-dimensional setting,
a special case of the multi-dimensional setting,
e.g. $\geq$$2$ feature-sequences per video \cite{shokoohi2015non}.
The term ``{\underline {dimension}}'' here can be understood as
a specific way of extracting feature-sequence from videos.
Our setting is multi-dimensional (Figure \ref{fig:sequentialisation}).
Specifically, defining a dimension in our context
is related to one of the two aspects:
(i) temporal pyramid ($h_t$ levels);
and (ii) spatial pyramid ($h_s$ levels);
Thus, we have a total of $h_t$$\times$$h_s$ dimensions
(feature extraction ways).
Note that, two feature-sequences at different dimensions
for the same video may have different lengths,
e.g. those extracted at different temporal pyramid levels
(Section \ref{sec:temopral-pyramids}).

Generally, there are two strategies to combine information
from multiple dimensions of sequences:
(1) {\em dependent}, and (2) {\em independent}.
We will generalise our TS-DTW model to the multi-dimensional setting using both strategies
as detailed below.

\paragraph{\bf (I) Dependent fusion}
The dependent fusion strategy assumes that:
(1) feature-sequences of a given video at different dimensions have the same length;
(2) different dimensions are strongly correlated one another,
i.e. their warping paths should be identical.
Due to condition (1), we can not perform fusion of multiple dimensions
across different temporal pyramid levels with this strategy.
Consequently, we can only combine the $h_s$
dimensions from different spatial divisions % and descriptor types
within each individual temporal pyramid level,
those extracted from the same slice-sequence.

Formally, when matching two slice-sequences of the same temporal pyramid level:
$S^p = \{s_1^p, \dots, s_i^p, \dots \}$ from video $Q^p$, and
$S^g = \{s_1^g, \dots, s_j^g, \dots \}$ from video $Q^g$,
we perform a joint sequence alignment
by using the feature data of all dimensions
to compute the distance between two elements $s_i^p$ and $s_j^g$ as
\begin{equation}
	\text{dist}_{\text{el}}^{\text{D}}(s_i^p,s_j^g) = \sum_{k=1}^{\kappa} \alpha_k \times \text{dist}_{\text{el}}(\mathbf{x}_{(i,k)}^p, \mathbf{x}_{(j,k)}^g)
	\label{eqn:dist_dependent}
\end{equation}
where $\mathbf{x}_{(i,k)}^p$ and $\mathbf{x}_{(j,k)}^g$
are the feature data in the $k$-th dimension for
$s_i^p$ and $s_j^g$, respectively,
$\kappa$ is the total number of dimensions to be fused, and
$\alpha_k$ defines the weight of the $k$-th dimension.
To incorporate the fine-to-coarse spatial information
encoded in walking motion,
we relate the value of $\alpha_k$ to the structure of spatial pyramid by
setting
\begin{equation}
	\alpha_k = 2^{\varepsilon_k} %\times \varphi
	\label{eqn:weight_dependent}
\end{equation}
where
$\varepsilon_k \in [0,1,\dots,h_s-1]$ denotes the spatial pyramid level of the $k$-th dimension
(see Figure \ref{fig:spatial-pyramid}).
This design is similar in spirit to
pyramid kernel matching \cite{grauman2007pyramid}.
All fused dimensions are at the same level of the temporal pyramid
whose structure is thus not considered here.

By replacing the single-dimensional distance
$\text{dist}_{\text{el}}(\cdot,\cdot)$ of DTW
with Eqn. (\ref{eqn:dist_dependent}),
our TS-DTW model can be readily generalised to
the multi-dimensional scenario and performs dimension fusion dependently.
We call this dependently generalised model ``{\bf MDTS-DTW$_\text{D}$}''.

\paragraph{\bf (II) Independent fusion}
In contrast to the dependent fusion policy,
the independent counterpart assumes independent alignment behaviours among
individual dimensions by performing information combination in the distance level.
Importantly, this strategy is more flexible than the former
as it allows each dimension having their respective sequence structure,
e.g. the sequence length.
Therefore, sequences across different temporal pyramid levels can be
combined in this fusion way.
Similarly, we further take into account temporal fine-to-coarse structures and
combine all dimensions to generate the final matching sequence distance
between two videos $Q^p$ and $Q^g$ via
\begin{equation}
	\text{dist}^\text{I}(Q^p, Q^g) = \sum_{k=1}^{\kappa}
	\beta_k \times \alpha_k \times \text{dist}_k(Q^p, Q^g)
	\label{eqn:dist_independent}
\end{equation}
where $\beta_k = 2^{\tau_k}, {\tau}_k \in \{0,1,\dots,h_t-1\}$
is the temporal pyramid level of the $k$-th dimension
(see Figure \ref{fig:sequentialisation}),
and $\text{dist}_k(Q^p, Q^g)$ the corresponding matching distance
using our TS-DTW model, i.e. Eqn. (\ref{eqn:dist_TSDTW}).
The parameters $\kappa$ and $\alpha_k$ are same as in
Eqns. (\ref{eqn:dist_dependent}) and (\ref{eqn:weight_dependent}).
We call this model ``{\bf MDTS-DTW$_\text{I}$}''

Usually, the two fusion strategies yield different matching results
over the same dimensions.
This is because each dimension may capture different aspects of video data
and produce non-identical alignment solutions,
and thus result in different distance values.
We will evaluate and discuss their performances for person ReID in Section \ref{sec:Exp}.

\subsection{Model Complexity}
\label{sec:model_complexity}
\xt{
We analyse the video matching complexity of our TS-DTW model.
Formally, given two feature-sequences $X^p$ and $X^g$,
we need to compute the matching distance between
$X^p(\Delta t)$ and $X^g(\Delta t)$ with
the time shift $\Delta t \in T=\{-|X^p|+1, \dots, |X^p|+|X^g|-1\}$.
$|X^p(\Delta t)|$ (or $|X^g(\Delta t)|$)
lies in the range of $[1, \text{min}(|X^p|, |X^g|)]$
(see Figure \ref{fig:time-shift}).
Therefore, the total matching complexity $\psi_\text{tsdtw}$ of
our TS-DTW model is
\begin{equation}
\label{eqn:model_complexity}
\psi_\text{tsdtw} =
\sum_{\Delta t \in T} \psi_\text{dtw} (|X^p(\Delta t)|)
\end{equation}
where $\psi_\text{dtw} (|X^p(\Delta t)|)$ refers to
the matching complexity of DTW,
which is $O(|X^p(\Delta t)|^2)$ by the standard DTW model \cite{berndt1994using}
and $O(|X^p(\Delta t)|)$ by fast variants \cite{salvador2004fastdtw}.
As person feature-sequences are typically short
(e.g. $<$$25$ on PRID$2011$ and $<$$40$ on iLIDS-VID),
the entire matching process is still efficient.
Moreover, we can parallelise easily the matching process over individual
time shifts for further reducing the running time,
as they are independent against each other.}

%--------------------------------------------------------------------
\section{Person Re-Identification}
\label{sec:ReID}

Given a probe person video $Q^p \in P$ and a gallery set $G = \{Q_i^g\}$
captured from two non-overlapping cameras,
person ReID aims to find the true identity match of ${Q}^p$ in $G$.
To achieve this, we first compute the space-time feature based distance
$\text{dist}^\text{st}(Q^p, Q_i^g)$
between $Q^p$ and every gallery video $Q_i^g$ with our
TS-DTW (Eqn. (\ref{eqn:dist_TSDTW})) or
MDTS-DTW$_\text{D}$ (Eqn. (\ref{eqn:dist_dependent})) or
MDTS-DTW$_\text{I}$ (Eqn. (\ref{eqn:dist_independent})) model.
In this way, we can obtain all cross-camera pairwise 
video matching distances
$\{ \text{dist}^\text{st}(Q^p, Q_i^g) \}_{i=1}^{|G|}$.
Finally, we generate a ranked list of all the gallery people
in ascendant order of their matching distances,
where the rank-$1$ gallery video is considered to be the most likely true match of $Q^p$.

\vspace{0.1cm}
\noindent \textbf{Combination with the spatial appearance methods.}
The ReID matching distances computed by the proposed model
can be readily fused with those by other spatial appearance models.
In particular, we incorporate our results $\text{dist}^\text{st}(Q^p, Q_i^g)$ into
other appearance based distance measures $\{\text{dist}_k^\text{sp}\}$ as
%the $i$-th spatial appearance method as
\begin{equation}
	{\text{dist}^\text{fused}}({Q^p}, Q_i^g) =
	\text{dist}^\text{st}(Q^p, Q_i^g) + \sum_k c_k  \times \text{dist}_k^\text{sp}({Q^p}, Q_i^g)
	\label{eqn:combination}
\end{equation}
where $c_i$ is a weighting assigned to the $k$-th method.
Instead of cross-validation, we simply set $c_k = 1$ for generality consideration
since in practice it is not always valid to assume the availability of pairwise labelled data which is required by cross-validation.
\xt{As matching distances by distinct methods may lie in different ranges,
we normalise all per-probe pairwise distances
$\text{dist}^\text{st}(Q^p, Q_i^g)/\text{dist}_k^\text{sp}({Q^p}, Q_i^g)$ 
to $[0, 1]$ per method separately
before performing fusion.
Specifically, given any matching distance $\text{dist}^* \in \{\text{dist}^\text{st}, \text{dist}_1^\text{sp}, \dots, \text{dist}_k^\text{sp}, \dots \}$,
we rescale all distances $\{\text{dist}^*(Q^p, Q_i^g)\}_{i=1}^{|G|}$ with respect to a probe $Q^p$ as 
\begin{equation}
\widehat{\text{dist}}^*({Q^p}, Q_i^g) =
\frac{\text{dist}^*(Q^p, Q_i^g)}
{\text{max}(\{\text{dist}^*(Q^p, Q_i^g)\}_{i=1}^{|G|})}
\label{eqn:norm_dist}
\end{equation}
where $\text{max}(\cdot)$ returns the maximal value of a set.
Then, the final fused distance can be expressed as
\begin{equation}
\widehat{\text{dist}}^\text{fused}({Q^p}, Q_i^g) =
\widehat{\text{dist}}^\text{st}(Q^p, Q_i^g) + \sum_k \widehat{\text{dist}}_k^\text{sp}({Q^p}, Q_i^g)
\label{eqn:combination_norm}
\end{equation}
We will evaluate the complementary effect between
space-time and appearance features based person ReID methods
in Section \ref{sec:Exp}.
}

%-------------------------------------------------------
\section{Experimental Settings}
\label{sec:Exp_Settings}

%sssssssssssssssssssssssssssssssss
\subsection{Datasets}
\label{sec:datasets}

Two benchmark image sequence based person ReID datasets
(PRID$2011$~\cite{hirzer11a} and
iLIDS-VID~\cite{wang2014person})
were utilised for evaluating the performance of the proposed approach.
Both datasets are challenging due to the large cross-view covariates
in view point, illumination condition, and background noises.
The dataset details are given below.

\begin{figure*} %[!htbp]
	\centering
	\subfigure %[PRID$2011$ \cite{hirzer11a}] %original]
	{
		\includegraphics[width=0.47\linewidth]{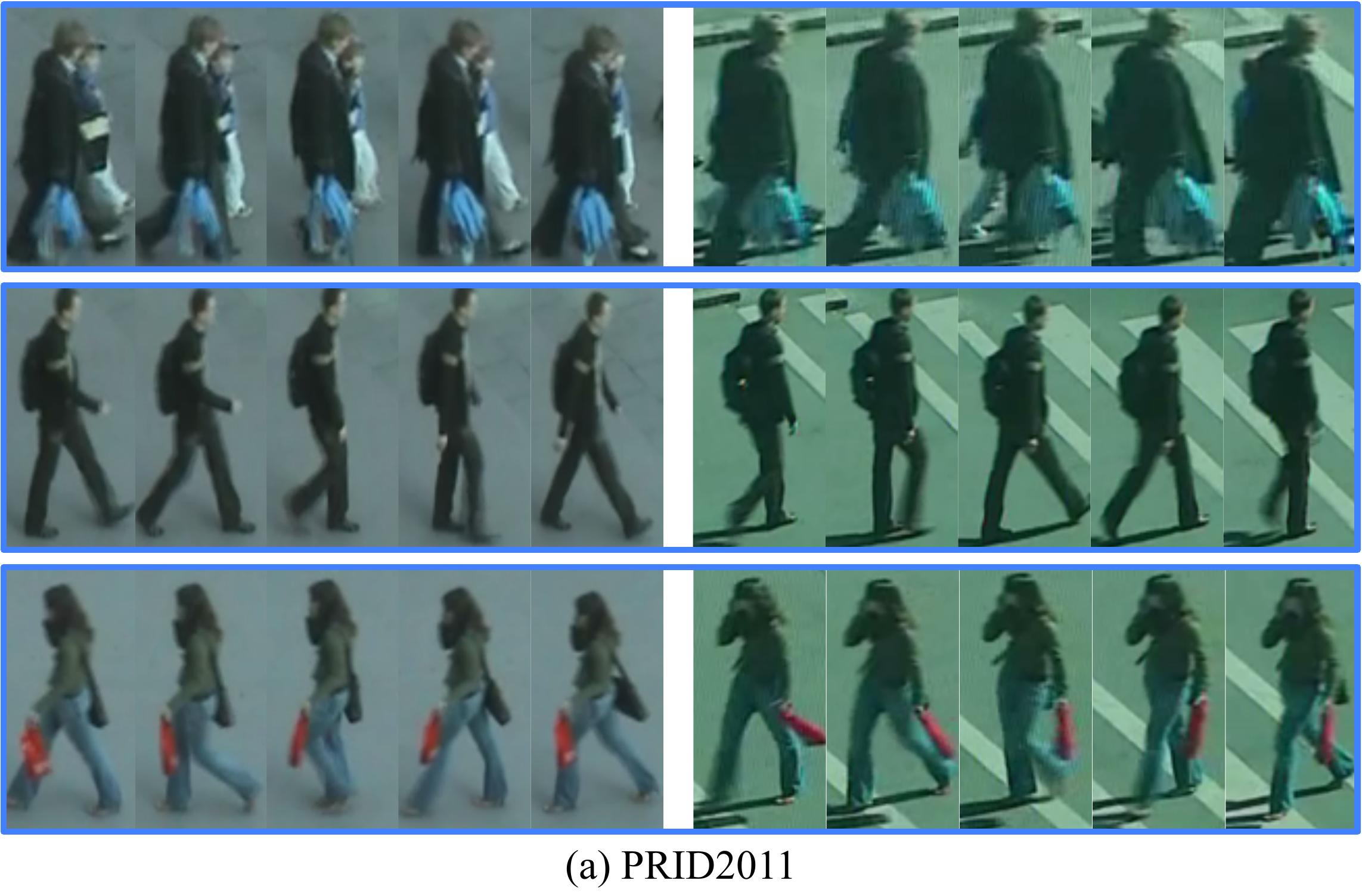}
	} \hfil
	\subfigure %[iLIDS-VID \cite{wang2014person}] %original]
	{
		\includegraphics[width=0.47\linewidth]{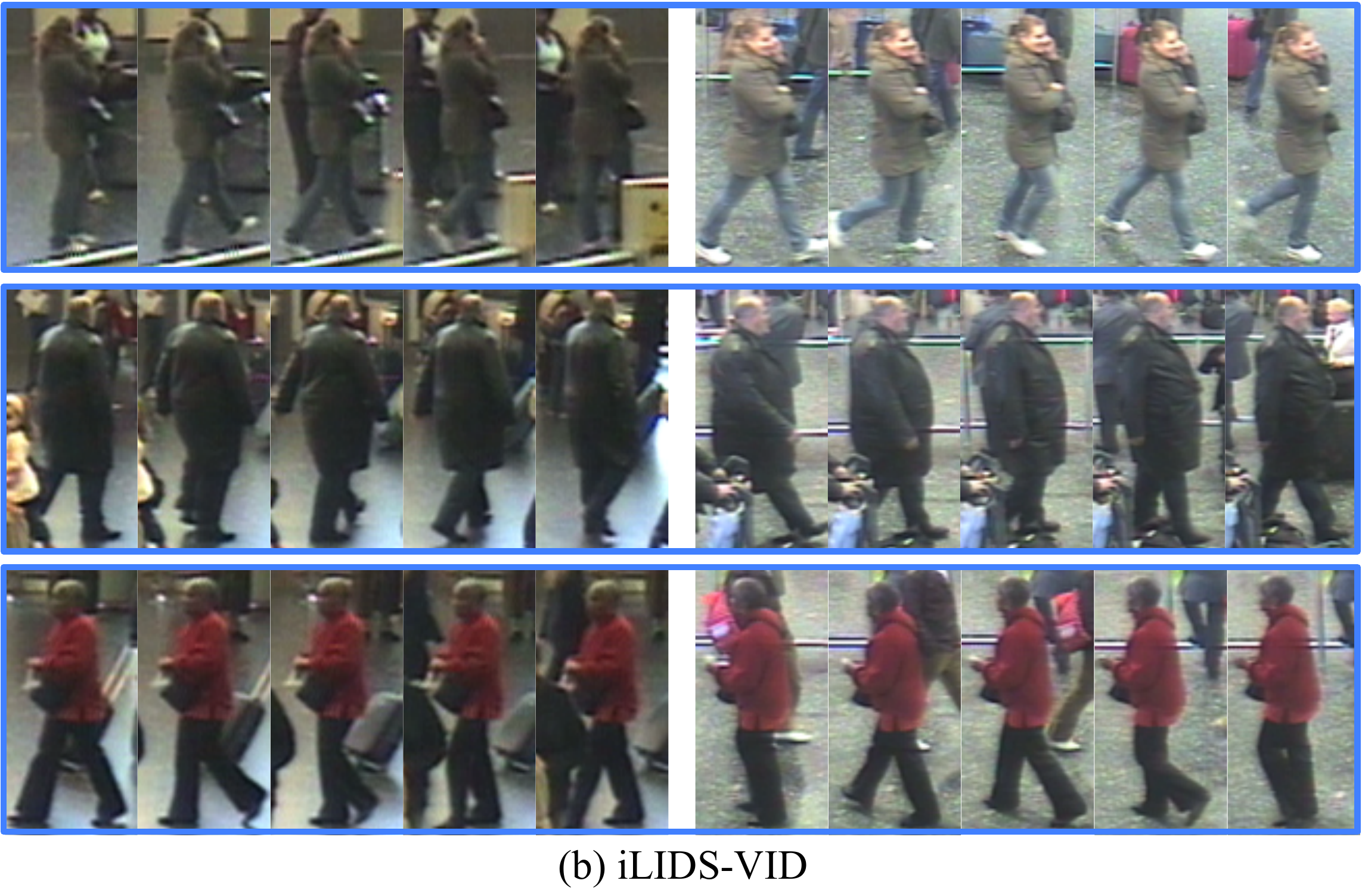}
	}
	\vskip -0.3cm
	\caption{
		\xt{Example person videos from the 
			(a) PRID$2011$ \cite{hirzer11a}
			and (b) iLIDS-VID \cite{wang2014person} 
			datasets.
		In each dataset, every blue bounding box contains two videos from the same person captured
		by two non-overlapping camera views.}
	}
	\label{fig:datasets}
\end{figure*}

\xt{
\begin{enumerate}
\item {\em PRID$2011$.}
The PRID$2011$ dataset~\cite{hirzer11a} includes $400$
image sequences captured from $200$ different people under
two disjoint outdoor camera views.
Each image sequence contains $5$ to $675$ image frames\footnote{For a fair comparison with existing methods, we followed the setting in \cite{wang2014person},
i.e. sequences of more than $21$ frames from $178$ people
were selected and utilised in our evaluations.} (Figure~\ref{fig:datasets}a).
\item {\em iLIDS-VID.}
The iLIDS-VID dataset~\cite{wang2014person}
contains a total of $600$ image sequences from $300$ randomly sampled
people, each with one pair of image sequences from two indoor camera views.
Every image sequence has a variable length, e.g. consisting of
$22$ to $192$ image frames \xt{(Figure~\ref{fig:datasets}b)}.
Compared with PRID$2011$, this dataset has
more complex occlusion and background clutter.
\end{enumerate}
}

%sssssssssssssssssssssssssssssssss
\subsection{Baseline Methods}
\label{sec:baselines}

We compared our method with related state-of-the-art methods as follows:
\begin{enumerate}
	
	%(1)
	\item \textit{GEI-RSVM}~\cite{martin2012gait}:
	A state-of-the-art gait recognition model
	using Gait Energy Image (GEI) feature~\cite{Han2006GEI} and
	the ranking SVM~\cite{chapelle2010efficient} model.
	\item \textit{DTW}~\cite{berndt1994using}:
	The widely used sequence matching algorithm - Dynamic Time Warping.
	DTW measures the distance between two sequences
	based on the optimal non-linear warping of elements
	across sequences.
	\item \textit{DDTW}~\cite{gullo2009time}:
	In contrast to DTW directly comparing feature values of elements
	that can be sensitive to diverse variations, DDTW considers
	the global shape of sequences by matching the first derivative of
	the original sequences.
	Besides, DDTW allows to avoid singularities,
	i.e. a single element of one sequence may map with a large partition
	of another sequence, which may lead to pathological measures \cite{keogh2001derivative}.
	\item \textit{WDTW} \cite{jeong2011weighted}:
	The weighted form of DTW model that also takes into account the shape similarity
	between two sequences.
	Specifically, WDTW introduces a multiplicative weight penalty
	on the warping distance between elements during distance estimation.
	This may suppress the negative influence of some outlier elements that
	are far away in element index but happen to be well matched.
	This model usually prefers close warping.
	We utilised a logistic weight function of the warping index-difference $\text{abs}(w_k^p - w_k^q)$ as:
	$f(w_k^p,w_k^q) = \frac{1}{1+\text{exp}(-(\text{abs}(w_k^p - w_k^q) - \mu)/2)}$,
	where $\mu$ is the half average-length of two sequences $Q^p$ and $Q^g$;
	$w_k^p$ and $w_k^q$ are the corresponding aligned element index
	of the $k$-th warp path entry (Eqn. (\ref{eqn:warp_path})).
	\item \textit{SDALF}~\cite{farenzena2010person}:
	A classic hand-crafted visual appearance ReID feature.
	Both single and multiple shot cases are considered.
	\item \textit{eSDC}~\cite{zhao2013unsupervised}:
	A state-of-the-art unsupervised spatial appearance based ReID method,
	which is able to learn localised appearance saliency statistics
	for measuring local patch importance.
	\xt
	{
	\item {\em Iterative Sparse Ranking} (ISR) \cite{lisanti2014person}:
	A contemporary weighted dictionary learning based algorithm that iteratively
	extends sparse discriminative classifiers in a transductive learning manner.
	}
	%
%	%
	\item {\em Regularised Dictionary Learning} (RDL) \cite{ElyorBMVC15}:
	The most recent dictionary learning based unsupervised ReID model.
	It iteratively learns the dictionary with the regularisation term updated in each iteration so that the cross-view noisy correspondence can be improved gradually.
	\item \textit{SS-ColLBP}~\cite{hirzer2012relaxed}:
	A ranking SVM model~\cite{chapelle2010efficient} based ReID method
	with one of the most effective features Colour$\&$LBP~\cite{hirzer2012relaxed}.
	\item \textit{MS-ColLBP}~\cite{wang2014person}:
	A multi-shot extension of SS-ColLBP.
	Specifically, the averaged Colour$\&$LBP feature~\cite{hirzer2012relaxed}
	over all image frames of a video
	is used to represent the spatial appearance of the person.
	\item \textit{$L_1$/$L_2$-norm}:
	The basic common distance metrics that can be very competitive with
	other complex metrics in many cases \cite{ding2008querying}.
	For matching two sequences, we remove the tail part of the longer one to
	make the two sequences have an equal duration.
	\xt
	{
	\item{\em Kernelised Cross-View Discriminant Component Analysis} (KCVDCA) \cite{chen2015CVDCA}:
	A competitive asymmetric distance learning method capable of inducing camera-specific projections for transforming unmatched visual features from different camera views to a shared subspace
	wherein discriminative features can be then learned and extracted.
	}
	\item \textit{Cross-View Quadratic Discriminant Analysis (XQDA)} \cite{liao2015person}:
	A state-of-the-art static appearance feature based supervised person ReID approach.
	Specifically,
%	the LOMO representation encodes and maximises the horizontal occurrence of local features for obtaining the viewpoint invariance capability.
	the XQDA algorithm learns simultaneously a discriminant low dimensional subspace and a QDA metric on the derived subspace.
	\item \textit{DVR}~\cite{WangDVRpami}:
	The state-of-the-art image-sequence based person ReID model
	which achieves the most competitive performance.
	In particular, this supervised model is characterised by discriminative fragment selection and exploitation
	for learning an effective space-time ranking function.

\end{enumerate}

%sssssssssssssssssssssssssssssssss
\subsection{Person ReID Scenarios}
\label{sec:reid_situations}
\xt{
We evaluated two person ReID scenarios,
closed-world and open-world:
\begin{enumerate}
\item {\em Closed-World ReID}:
In this setting, all probe people are assumed to exist in the gallery.
In evaluations, we followed the data partition setting as \cite{wang2014person,WangDVRpami}.
Specifically, for either PRID$2011$ or iLIDS-VID,
we split the entire dataset into two partitions:
one half for training, and the other half for testing.
Note that our model does not utilise the training partition since it is unsupervised.
%
%\vspace{0.1cm}
%\noindent {\bf Open-World ReID.}
\item {\em Open-World ReID}:
In addition, we evaluated a more realistic scenario
called open-world ReID~\cite{liao2014open}.
Specifically, its key difference from the closed-world case is that
a probe person $i \in P$ is not assumed to appear {\em necessarily}
in the gallery $G$ under the open-world setting.
This situation is more plausible to real-world ReID applications
since we generally have no prior knowledge about
whether one person (in gallery) re-appears in
certain (probe) camera views in most applications,
e.g. due to the complex topology structure of camera networks.
That is, $P$ and $G$ may be just partially overlapped in different
camera views.
Similar data partitions as the closed-world case were utilised,
with the only difference that the gallery set of the testing partition
is reduced by one third ($\frac{1}{3}$) of randomly selected people
(they are considered as imposters, only appearing in the probe set),
i.e. $60$ gallery people on PRID$2011$ and $100$ on iLIDS-VID.
% Similarly, we reported the averaged results over $10$ folds of experiment.
\end{enumerate}
}

%sssssssssssssssssssssssssssssssss
\subsection{\xt{Evaluation Metrics}}
\label{sec:eval_metrics}
\xt{
For closed-world ReID,
the conventional Cumulated Matching
Characteristics (CMC) curves were utilised for a quantitative
performance comparison between different methods \cite{gong2014person}.
%
%
%\vspace{0.1cm}
%\noindent {\bf Open-World ReID.}
For open-world ReID,
two separate steps are involved in performance evaluation
under the open-world setting~\cite{liao2014open}:
(1) Detection - decide if a probe person $Q^p \in P$ exists in the gallery or not;
For convenience, we define $\bar{P} = P \setminus G$, the probe people that are not included in the gallery $G$.
(2) Identification - compute the truly matched rates over only {\em accepted} target people.
Specifically, we utilised detection and identification rate (DIR)
and false accept rate (FAR) defined as:
\begin{equation} \small
\text{DIR}(\tau, k)=\frac
{|\{Q^p| \hat{Q}^g \in G,\text{rank}(Q^p) \leq k,\ \text{dist}(\hat{Q}^g, Q^p) \leq \tau \} |}{|G|}
\label{eqn:DIR}
\end{equation}
%
%\vspace{-.2cm}
\begin{equation} \small
\text{FAR}(\tau)=\frac
{|\{Q^p| Q^p \in \bar{P}, \min_{Q^g \in G} \text{dist}(Q^g,Q^p) \leq \tau \} |}{|\bar{P}|}
\label{eqn:FAR}
\end{equation}
%\vspace{-.cm}
%
where $\text{dist}(\cdot,\cdot)$ refers to
the cross-view distance score induced by some person ReID model,
$\hat{Q}^g$ the gallery person having the same identity (i.e. true match)
as the probe person $Q^p$,
and $\tau$ the decision threshold.
$\text{rank}(\hat{Q}^g)=k$ means that the true match $\hat{Q}^g$ is ranked at $k$ in the ranking list.
Thus, given a rank $k$, a Receiver Operating Characteristic (ROC) curve
can be obtained by varying $\tau$.
}

%\vspace{0.1cm}
%\noindent {\bf Parameter settings.}
%sssssssssssssssssssssssssssssssss
\subsection{Implementation Details}

Since video slices are localised over time,
the value of $l$ (the shortest slice length) should be small
and related to the walking cycle length.
We fixed $l=5$ in that the process of a walking step takes around $2l = 10$ frames.
Whilst the size $h_t$ of the temporal pyramid largely depends on video length,
e.g. an over-large $h_t$ may lead to discarding many frames during sequentialisation
(thus causing potentially much information loss), or
very few slices produced for videos
(with little temporal ordering dynamics).
Thus, $h_t$ is set to $2$ accordingly.
We utilised a $2$-level spatial pyramid, i.e. $h_s = 2$.
This is because,
our empirical experiments suggest that the addition of one more spatial pyramid level
slightly degrades the model performance
possibly due to the local patch misalignment problem
in over fine-grained spatial decomposition.
The distance metric between sequence elements $\text{dist}_{\text{el}}(\cdot,\cdot)$ is set as $L_1$.

For obtaining stable statistics,
we evaluated both person ReID scenarios with $10$ folds of experiments
with different random training/testing partitions on each dataset,
and reported the averaged results.

%------------------------------------------------------
\section{Experimental Results}
\label{sec:Exp}

\begin{table} %[h]
	\renewcommand{\arraystretch}{1}
	\centering
	\footnotesize %
	%\scriptsize
	\setlength{\tabcolsep}{.05cm}
	\caption{
		The \xt{\em closed-world} person ReID performance of the proposed TS-DTW (single-dimensional)
		and MDTS-DTW (multi-dimensional) model with different parts of our STPS video representation.
		(TPL: Temporal Pyramid Level;
		SPL: Spatial Pyramid Level)
	}
	%\vskip -0.3cm
	\label{tab:model_analysis}
	%\vspace{0.2cm}
	\begin{tabular}{l|cccc|cccc}
		\hline%\noalign{\smallskip}
		Dataset
		& \multicolumn{4}{c|}{PRID$2011$ \cite{hirzer11a}}
		& \multicolumn{4}{c}{iLIDS-VID \cite{wang2014person}}
		\\
		\hline%\noalign{\smallskip}
		Rank $R$ (\%)
		& 1 & 5 & 10 & 20
		& 1 & 5 & 10 & 20
		%& 1 & 5 & 10 & 20 
		\\%
		\hline
		\hline
		TS-DTW(TPL$^0$,SPL$^0$) %-fl5numofcells+HE(0:1)
		&{36.7}&{59.1}&{73.5}&{84.7}
		&{23.3}&{51.5}&{65.2}&{79.6}
		%&34.4&61.4&67.2&81.0
		\\
		\hline
		TS-DTW(TPL$^0$,SPL$^1$) %-fl5numofcells+HE(1:0)
		&{32.5}&{63.8}&{75.4}&{84.9}
		&{12.3}&{37.0}&{53.2}&{68.5}
		%&23.4&46.4&61.4&79.0
		\\
		\hline
		MDTS-DTW$_\text{D}$(TPL$^0$)
		&{37.1}&{60.2}&{73.7}&{85.7}
		&{25.1}&{51.9}&{66.5}&{79.9}
		%&35.8&61.6&66.8&82.0 
		\\
		\hline
		MDTS-DTW$_\text{I}$(TPL$^0$) %-fl5numofcells+HE(1:2)
		&{39.2}&{60.8}&{75.3}&{86.6}
		&{25.9}&{52.7}&{67.1}&{79.1}
		%&{\bf 40.6}&{\bf 63.1}&{\bf 71.7}&{\bf 85.7}
		\\
		\hline
		\hline
		TS-DTW(TPL$^1$,SPL$^0$) %-fl10numofcells+HE(0:1)
		&{34.2}&{58.9}&{74.4}&{86.1}
		&{23.8}&{49.5}&{62.7}&{78.4}
		%&-&-&-&-
		\\
		\hline
		TS-DTW(TPL$^1$,SPL$^1$) %-fl10numofcells+HE(1:0)
		&{32.4}&{61.7}&{77.0}&{87.2}
		&{16.5}&{40.7}&{53.4}&{68.7}
		%&-&-&-&-
		\\
		\hline
		MDTS-DTW$_\text{D}$(TPL$^1$)
		&{36.2}&{60.3}&{74.8}&{86.3}
		&{23.8}&{50.0}&{62.5}&{78.6}
		%&-&-&-&- 
		\\
		\hline
		MDTS-DTW$_\text{I}$(TPL$^1$) %-fl10numofcells+HE(1:2)
		&{37.2}&{61.7}&{75.2}&{87.0}
		&{24.3}&{50.1}&{62.4}&{78.5}
		%&-&-&-&-
		\\
		\hline
		\hline
		MDTS-DTW$_\text{I}$(full) %-fl5(2)+10(1)numofcells(1:2)+HE
		&{\bf 41.7}&{\bf 67.1}&{\bf 79.4}&{\bf 90.1}
		&{\bf 31.5}&{\bf 62.1}&{\bf 72.8}&{\bf 82.4}
		%&-&-&-&-
		\\
		\hline
	\end{tabular}
	%\vspace{-0.2cm}
\end{table}

\subsection{Evaluation on Our Proposed Approach}
\label{sec:Eval_our_model}
%************************************************
% \subsubsection{Effect of Spatio-Temporal Pyramids}
We evaluated
the detailed aspects of the proposed video representation
and sequence matching models for person ReID
in the common {\em closed-world} scenario,
i.e. the ReID accuracies of our TS-DTW and MDTS-DTW models
using different parts of the proposed STPS features.
The results are reported in Table \ref{tab:model_analysis}.
It is evident that both temporal and spatial pyramids
are effective for person ReID and their fusion with the proposed method can
improve significantly the matching accuracy.
This is consistent with the finding
in scene and action recognition \cite{lazebnik2006beyond,pirsiavash2012detecting}.

Specifically, given either of the two temporal pyramid levels,
when comparing with the coarse spatial pyramid level (SPL-$1$),
the fine-grained spatial division (SPL-$0$) produces similar result on PRID$2011$,
but significantly better accuracy on the more challenging iLIDS-VID.
In contrast, with the same SPL,
two temporal pyramid levels (TPL-$0$ and TPL-$1$) produce similar results.
The plausible reason is that larger spatial regions are
more likely to be contaminated by random noise in a crowded public space.
%=============
% fusion
When combining the matching results from different
dimensions/feature-sequences of the same temporal pyramid level by either MDTS-DTW$_\text{D}$ or MDTS-DTW$_\text{I}$, the ReID accuracy can be improved similarly on both datasets.
This suggests largely the independence property among distinct sequence dimensions,
i.e. modelling their dependence does not bring any benefit in enhancing ReID.
Moreover, after the results from different temporal granularities are fused by MDTS-DTW$_\text{I}$,
ReID accuracies are further increased
(note, MDTS-DTW$_\text{D}$ is not able to fuse image sequences of different lengths, see Section \ref{sec:multi-dimension}).
These evidences show good complementary effect of different spatio-temporal pyramid levels
and effectiveness of our model in fusing information
from multiple localised motion patterns with different space-time extends.
In the remaining evaluations, we utilised our MDTS-DTW$_\text{I}$ model and the full STPS video representation for comparison with the baseline methods.

\xt{
{\em Computational cost}:
Apart from person re-id accuracy,
we also evaluated the computational cost of
our MDTS-DTW$_\text{I}$ model on matching 
cross-view person videos for ReID.
Time was measured on a work station
(Intel i$7$-$4770$K CPU at $3.50$ GHz and memory of $16$ GB)
with Matlab implementation in Windows OS. % of Version R2013a.
Time analysis was conducted % on PRID and iLIDS-VID 
under the same experimental setting as above.
%Specifically, a total of $89$ (PRID) or 150 (iLIDS-VID) probe people were performed.
On average, matching each probe video against the
gallery set takes 
$5.26$ % $1.756 * 3$ 
seconds on PRID ($89$ gallery people)
and 
$9.50$ %$3.179$ * 3 
seconds on iLIDS-VID ($150$ gallery people).
That is, the average matching time for two
person sequences is around $0.06$ second.
Note that,
the whole process above can be conducted in parallel 
over a cluster of machines
to further speed up model deployment.
}

%---------------------------------------------------------
\subsection{Evaluation on Closed-World Person ReID}

In this conventional setting, we performed comparative
evaluations with gait recognition, temporal sequence matching, and person ReID approaches.

%------------------------------------------------------
\begin{table} %[h] %[!htbp]
	\renewcommand{\arraystretch}{1}
	\centering
	\footnotesize %
	%\scriptsize
	\setlength{\tabcolsep}{.1cm}
	\caption{
		Comparing gait recognition and sequence matching methods
		\xt{(closed-world scenario).}
	}
	%\vskip -0.3cm
	\label{tab:gait_sequence_match}
	\begin{tabular}{l|cccc|cccc}
		\hline
		Dataset
		& \multicolumn{4}{c|}{PRID$2011$ \cite{hirzer11a}}
		& \multicolumn{4}{c}{iLIDS-VID \cite{wang2014person}}
		\\
		\hline
		Rank $R$ (\%)
		& 1 & 5 & 10 & 20
		& 1 & 5 & 10 & 20
		%& 1 & 5 & 10 & 20
		\\
		\hline \hline
		GEI-RSVM \cite{martin2012gait}
		& 20.9 & 45.5 & 58.3 &70.9
		&	2.8		&	13.1	&	21.3	&	34.5
		%&3.0&17.6&33.0&54.4 
		\\
		\hline
		DTW \cite{berndt1994using}
		&19.9&41.2&53.6&65.8
		&15.9&32.1&41.5&55.5
		%&14.6&28.6&44.9&74.9
		\\
		% *** using our space-time pyramid ***
		\hline
		% DDTW$_\text{D}$
		DDTW \cite{gullo2009time}
		& 5.4 & 18.2 & 27.5 & 38.5
		& 2.9 & 10.1 & 18.1 & 31.5
		%&5.7&20.3&35.4&64.0
		\\
		\hline
		% WDTW$_\text{D}$
		WDTW \cite{jeong2011weighted}
		& 4.2 & 13.7 & 20.9 & 29.4
		& 5.1 & 11.5 & 16.0 & 23.9
		%&9.7&24.6&41.1&69.7
		\\
		\hline \hline
		{\bf MDTS-DTW$_\text{I}$} 
		&{\bf 41.7}&{\bf 67.1}&{\bf 79.4}&{\bf 90.1}
		&{\bf 31.5}&{\bf 62.1}&{\bf 72.8}&{\bf 82.4}
        %&{\bf 40.6}&{\bf 63.1}&{\bf 71.7}&{\bf 85.7} 
        \\
		\hline
	\end{tabular}
	%\vspace{-0.2cm}
\end{table}

%************************************************
\subsubsection{Comparing Gait Recognition and Temporal Sequence Matching Methods}

In Table \ref{tab:gait_sequence_match}, we compared our MDTS-DTW$_\text{I}$ model with a number of state-of-the-art gait recognition and
dynamic programming based sequence matching methods.
It is evident that the proposed model
outperforms both alternative strategies by a large margin on each dataset.
Specifically, the gait recognition method produces much better ReID accuracy on PRID2011
than on iLIDS-VID. This is because, the image sequences from the latter contain
more background noise such as clutter and occlusion which can
contaminate the gait feature heavily (see Figure \ref{fig:GEI}).
By automatically aligning starting/ending walking phases and selecting best-matched
sequence parts, our TS-DTW model allows to better overcome this challenge.
On the other hand, conventional temporal sequence matching algorithms,
e.g. DTW and its variants,
can only provide much weaker results than the proposed MDTS-DTW.
%Particularly, the DDTW model exploits sequence global shape information, and the WDTW algorithm enforces non-linear penalty on further alignment across sequences.
This is largely owing to:
(1) ReID image sequences have different lengths with arbitrary starting/ending phases, and incomplete/noisy frames. Hence, attempts to match and utilise entire sequences inevitably suffer from mismatching with erroneous similarity measurement;
(2) there is no explicit mechanism to avoid incomplete/missing data, typical in crowded surveillance scenes.

%************************************************
\subsubsection{Comparing Person ReID Methods}

We compared our MDTS-DTW$_\text{I}$ method with contemporary unsupervised and supervised ReID methods, and further evaluated the complementary effect between appearance and space-time feature based approaches.

%------------------------------------------------------
\begin{table} %[h] %[!htbp]
	\renewcommand{\arraystretch}{1}
	\centering
	\footnotesize %
	%\scriptsize
	\setlength{\tabcolsep}{.1cm}
	\caption{
		Comparing unsupervised person ReID methods
		\xt{(closed-world scenario).}	
	}
	%\vskip -0.3cm
	\label{tab:unsupervised}
	\begin{tabular}{l|cccc|cccc}
		\hline%\noalign{\smallskip}
		Dataset
		& \multicolumn{4}{c|}{PRID$2011$ \cite{hirzer11a}}
		& \multicolumn{4}{c}{iLIDS-VID \cite{wang2014person}}
		\\
		\hline%\noalign{\smallskip}
		Rank $R$ (\%)
		& 1 & 5 & 10 & 20
		& 1 & 5 & 10 & 20
		%& 1 & 5 & 10 & 20
		\\
		\hline \hline%\noalign{\smallskip}
		$L_1$-norm			
		& 26.4&47.5&57.8&73.7
		& 19.3&39.2&51.9&66.5
		%&28.6&50.9&67.1&85.7 
		\\
		$L_2$-norm
		& 23.3&46.7&57.5&73.6
		& 15.6&37.7&49.0&63.1
		%&18.9&47.7&69.1&86.9  
		\\
		\hline
		SS-SDALF \cite{farenzena2010person}	
		&	4.9	&	21.5	&	30.9	&	45.2
		&	5.1	&	14.9	&	20.7	&	31.3
		%&24.6&55.7&73.7&85.4 
		\\
		MS-SDALF \cite{farenzena2010person}	
		&	5.2	&	20.7	&	32.0	&	47.9
		&	6.3	&	18.8	&	27.1	&	37.3
		%&15.4&50.6&76.0&87.7
		\\
		\hline
		\xt{ISR} \cite{lisanti2014person}
		& 17.3 & 38.2 & 53.4 & 64.5
		& 7.9 & 22.8 & 30.3 & 41.8
		%&11.1&44.9&60.0&78.9
		\\
		\hline
		eSDC \cite{zhao2013unsupervised}
		&	25.8	&	43.6	&	52.6	&	62.0
		&	10.2	&	24.8	&	35.5	&	52.9
		%&22.0&54.6&{\bf 78.2}&{\bf 94.0}
		\\
		\hline
		RDL \cite{ElyorBMVC15}
		& 29.1 & 53.6 & 66.2 & 76.1
		& 11.5 & 26.2 & 34.3 & 46.3
		%&18.6&44.0&62.0&79.8    %26.0&59.7&76.3&93.7
		\\
		\hline
		\textbf{MDTS-DTW$_\text{I}$}
		&{\bf 41.7}&{\bf 67.1}&{\bf 79.4}&{\bf 90.1}
		&{\bf 31.5}&{\bf 62.1}&{\bf 72.8}&{\bf 82.4}
		%&{\bf 40.6}&{\bf 63.1}&71.7&85.7 
		\\
		\hline
	\end{tabular}
\end{table}

\vspace{0.1cm}
\noindent {\bf Comparing unsupervised methods}.
Table \ref{tab:unsupervised} shows the comparison among unsupervised ReID approaches.
The proposed MDTS-DTW$_\text{I}$ outperforms significantly all competitors on PRID2011 and iLIDS-VID. Specifically, space-time feature based methods (e.g. ours and L$_1$/L$_2$-norm) produce better ReID accuracies than the remaining spatial appearance based methods,
% (e.g. SC, eSDC and SDALF),
particularly on the more challenging iLIDS-VID dataset.
This suggests the inherent challenge caused by the ambiguous and unreliable nature of people's appearance in person ReID applications,
and simultaneously the exceptional effectiveness of space-time cues for people matching
when expressed and exploited effectively.
In addition, the weak performance by SDALF is largely because of the
intrinsic difficulty in designing general identity-discriminative
hand-crafted appearance feature given
unknown cross-camera covariates.
% While eSDC and RDL improve the recognition rates
\xt{Through iteratively learning and extending discriminative classifiers in ISR or modelling localised saliency statistics in eSDC or exploiting iteratively cross-view soft-correspondence in RDL, 
	person ReID performance is greatly improved.}
However, due to relying on static appearance information alone,
they are inherently sensitive to cross-camera viewing conditions,
e.g. with a severe perform degradation from PRID$2011$ to iLIDS-VID.
In contrast, our method mitigates this challenge
by properly designing and effectively exploiting dynamic space-time features,
another information source which presents better stability than the widely-used appearance features.

\begin{table} [h] %[!htbp]
	%\vspace{-0.1cm}
	\renewcommand{\arraystretch}{1}
	\setlength{\tabcolsep}{.15cm}
	\centering
	\footnotesize %
	%\scriptsize
	\setlength{\tabcolsep}{.1cm}
	\caption{
		Comparing supervised person ReID methods
		\xt{(closed-world scenario).}
	}
	%\vskip -0.3cm
	\label{tab:supervised}
	\begin{tabular}{l|cccc|cccc}
		\hline
		Dataset
		& \multicolumn{4}{c|}{PRID$2011$ \cite{hirzer11a}}
		& \multicolumn{4}{c}{iLIDS-VID \cite{wang2014person}}
		\\
		\hline
		Rank $R$ (\%)
		& 1 & 5 & 10 & 20
		& 1 & 5 & 10 & 20
		%& 1 & 5 & 10 & 20
		\\
		\hline \hline					
		SS-ColLBP \cite{hirzer2012relaxed}
		&	22.4	&	41.8	&	51.0	&	64.7
		&	9.1	    &	22.6	&	33.2	&	45.5
		%&15.2&44.0&71.8&89.4 
		\\
		\hline
		MS-ColLBP \cite{hirzer2012relaxed}
		&{34.3}	&{56.0}	&{65.5}	&	77.3
		&{23.2}	&{44.2}	&	54.1	&{68.8}
		%&26.8&65.2&83.4&92.0 
		\\
		\hline
		DVR \cite{WangDVRpami}	
		& 40.0 & 71.7 & 84.5 & 92.2
		& {\bf 39.5} & {61.1} & {71.7} & {81.0}
		%&15.4&43.2&59.8&76.8 
		\\
		\hline
		\xt{KCVDCA} \cite{chen2015CVDCA}
		& 43.8 & 69.7 & 76.4 & 87.6
		& 16.7 & 43.3 & 54.0 & 70.7
		%&-&-&-&-
		\\ 
		\hline
		{XQDA} \cite{liao2015person}
		%& {\bf 48.0} & {\bf 79.2} & {\bf 89.3} & {\bf 96.5}
		& {\bf 46.3} & {\bf 78.2} & {\bf 89.1} & {\bf 96.3}
		& 16.7 & 39.1 & 52.3 & 66.8
		%&{\bf 48.4}&{\bf 85.2}&{\bf 96.2}&{\bf 98.8}
		\\ 
		\hline
		{\bf MDTS-DTW$_\text{I}$} %
		&{41.7}&{67.1}&{79.4}&{90.1}
		&{31.5}&{\bf 62.1}&{\bf 72.8}&{\bf 82.4}
		%&40.6&63.1&71.7&85.7 
		\\
		\hline
	\end{tabular}
	%\vspace{-0.2cm}
\end{table}

\begin{table*} %[h] %[!htbp]
	%\vspace{-0cm}
	\renewcommand{\arraystretch}{1}
	\setlength{\tabcolsep}{.15cm}
	\centering
	\footnotesize %
	%\scriptsize
	\setlength{\tabcolsep}{.35cm}
	\caption{
		Evaluating complementary effect 
		% Complementary 
		between space-time and appearance feature based person ReID methods
		\xt{(closed-world scenario).}
	}
	%\vskip -0.3cm
	\label{tab:combined}
	\vspace{0.2cm}
	\begin{tabular}{l|cccc|cccc}
		\hline
		Dataset
		& \multicolumn{4}{c|}{PRID$2011$ \cite{hirzer11a}}
		& \multicolumn{4}{c}{iLIDS-VID \cite{wang2014person}}
		\\
		\hline
		Rank $R$ (\%)
		& 1 & 5 & 10 & 20
		& 1 & 5 & 10 & 20
		%& 1 & 5 & 10 & 20
		\\
		\hline \hline
		DVR \cite{WangDVRpami}
		& 40.0 & 71.7 & 84.5 & 92.2
		& {39.5} & {61.1} & {71.7} & {81.0}
		% &15.4&43.2&59.8&76.8 
		\\
		{\bf MDTS-DTW$_\text{I}$}
		&{41.7}&{67.1}&{79.4}&{90.1}
		&{31.5}&{62.1}&{72.8}&{82.4}
		%&40.6&63.1&71.7&85.7
		\\
		\hline %======================
		eSDC\cite{zhao2013unsupervised}
		&	25.8	&	43.6	&	52.6	&	62.0
		&	10.2	&	24.8	&	35.5	&	52.9
		% &22.0&54.6&78.2&94.0
		\\
		eSDC+DVR\cite{WangDVRpami}
		& 44.3 & 68.4 & 78.2 & 91.1
		& 29.5 & 54.0 & 66.4 & 78.4
		% &31.6&67.0&82.2&93.0
		\\
		eSDC+{\bf MDTS-DTW$_\text{I}$}
		& 48.0 & 69.9 & 82.0 & 91.8
		& 33.5 & 64.1 & 74.2 & 83.5
		% &43.6&76.0&84.6&94.2
		\\
		\hline %======================
		\xt{ISR} \cite{lisanti2014person}
		& 17.3 & 38.2 & 53.4 & 64.5
		& 7.9 & 22.8 & 30.3 & 41.8
		% &11.1&44.9&60.0&78.9
		\\
		\xt{ISR+DVR}
		&43.8&63.3&72.5&81.3
		&30.0&46.0&55.1&63.6
		% &34.9&58.0&73.1&89.7
		\\
		\xt{ISR+{\bf MDTS-DTW$_\text{I}$}}
		&46.2&66.7&72.6&83.3
		&33.1&51.5&58.7&69.7
		% &40.3&57.4&77.4&87.7
		\\
		\hline
		RDL \cite{ElyorBMVC15}
		& 29.1 & 53.6 & 66.2 & 76.1
		& 11.5 & 26.2 & 34.3 & 46.3
		% &18.6&44.0&62.0&79.8 
		\\
		RDL+DVR
		& 58.9 & 79.7 & 87.5 & 93.6
		& 31.7 & 56.9 & 67.7 & 80.5
		% &18.6&44.0&62.0&79.8 
		\\
		RDL+{\bf MDTS-DTW$_\text{I}$}
		& 59.2 & 82.7 & 88.4 & 94.9
		& 35.3 & 63.4 & 73.9 & 83.3
		% &31.6&60.4&66.6&77.6
		\\
		\hline \hline
		MS-ColLBP \cite{hirzer2012relaxed}
		&	34.3	&	56.0	&	{65.5}&	77.3
		&	{23.2}&	{44.2}&	{54.1}&	{68.8}
		% &26.8&65.2&83.4&92.0
		\\
		MS-ColLBP+DVR
		& 44.8 & 66.9 & 77.1 & 89.9
		& 39.5 & 61.0 & 72.7 & 82.8
		% &48.2&72.2&84.2&94.2
		\\
		MS-ColLBP+\textbf{MDTS-DTW$_\text{I}$}	
		& 47.8 & 67.5 & 79.9 & 91.0
		& 44.1 & 69.9 & 79.1 & 88.8
		% &48.2&72.2&84.2&94.2
		\\		
		\hline %======================
		\xt{KCVDCA} \cite{chen2015CVDCA}
		& 43.8 & 69.7 & 76.4 & 87.6
		& 16.7 & 43.3 & 54.0 & 70.7
		% &-&-&-&-
		\\
		\xt{KCVDCA+DVR}
		&65.7&88.1&93.4&97.3
		&{\bf 54.9}&76.8&83.7&91.3
		% &-&-&-&-
		\\
		\xt{KCVDCA+{\bf MDTS-DTW$_\text{I}$}}
		&71.0&89.0&93.8&97.5%41.8&67.9&80.2&90.6
		&50.6&{\bf 77.0}&{\bf 85.6}&{\bf 92.6}%33.0&63.4&74.1&83.2
		% &-&-&-&-
		\\
		\hline %======================
		{XQDA} \cite{liao2015person}
		& {46.3} & {78.2} & {89.1} & {96.3}
		& 16.7 & 39.1 & 52.3 & 66.8
		% &48.4&{\bf 85.2}&{\bf 96.2}&{\bf 98.8}
		\\
		XQDA+DVR
		& {\bf 77.4} & {\bf 93.9} & {\bf 97.0} & {\bf 99.4}
		& 51.1 & 75.7 & {83.9} & {90.5}
		% &40.2&73.6&86.4&96.4
		\\
		XQDA+{\bf MDTS-DTW$_\text{I}$}
		& 69.6 & 89.4 & 94.3 & 97.9
		& 49.5 & 75.7 & 84.5 & 91.9
		% &{\bf 55.6}&76.6&86.6&95.8
		\\
		\hline
	\end{tabular}
	%\vspace{-0.4cm}
\end{table*}

\vspace{0.1cm}
\noindent {\bf Comparing supervised methods}.
We present the comparison between our unsupervised MDTS-DTW$_\text{I}$
and previous supervised methods in Table \ref{tab:supervised}.
It is found that space-time feature based methods (i.e. DVR \& ours)
are less sensitive to crowded background
than other appearance feature based models particularly XQDA and KCVDCA, when
comparing the ReID performance on PRID and iLIDS-VID
(more busy and crowded,
see Figure \ref{fig:datasets}).
This is partially attributed to the selective matching strategy in the former models for extracting more reliable space-time representations.
Moreover, it is observed that our method surpasses appearance based SS-/MS-ColLBP on two datasets
and XQDA/KCVDCA on iLIDS-VID, and produces competitive results as video based DVR.
Note that the DVR model exploits both space-time and colour information in the price of exhaustive pairwise labelling
whilst our MDTS-DTW$_\text{I}$ method only utilises dynamic space-time cues without the need for cross-view pairwise labelling.
These comparisons demonstrate the advantage and capability of our STPS video representation and selective matching model
in extracting and exploiting identity-discriminative space-time information from noisy person videos for relaxing the label availability assumption and making better use of unregulated video data.

\vspace{0.1cm}
\noindent {\bf Evaluating complementary effect}.
We further evaluated how well spatial appearance and space-time feature based ReID methods complement each other. To this end, we integrated contemporary unsupervised (eSDC, ISR and RDL) and supervised (MS-ColLBP, KCVDCA and XQDA)
appearance based approaches with DVR and our MDTS-DTW$_\text{I}$ model (Eqn. \eqref{eqn:combination}), respectively.
The results are presented in Table \ref{tab:combined}.
It is observed that by fusing space-time feature based ReID results of either DVR or ours,
the matching accuracies of existing appearance based methods can be significantly boosted.
This confirms the similar finding by \cite{WangDVRpami}
that, the combination of appearance and space-time motion information sources can be very effective for person ReID
as they are largely independent in nature.
\xt{Overall, XQDA+DVR achieves the best performance on PRID$2011$ whilst
KCVDCA+Ours and KCVDCA+DVR perform similarly best on iLIDS-VID.}
This is as expected because the combination with DVR 
doubly benefits much from
effective modelling on labelled data which
contain strong discriminative information but very expensive to acquire for
every camera pair in reality.
Once removing the label availability assumption,
the best results are obtained by eSDC+Ours on iLIDS-VID and RDL+Ours on PRID$2011$.
Under the unsupervised setting,
we observed a similar complementary effect as XQDA/KCVDCA+DVR/Ours.
This validates the efficacy of our ReID method in deriving dynamic
identity information from unregulated videos,
independent of and completing well the commonly used spatial appearance.

\begin{table*} %[h] %[!htbp]
	\renewcommand{\arraystretch}{1}
	\setlength{\tabcolsep}{.15cm}
	\centering
	\footnotesize
	%\scriptsize
	\setlength{\tabcolsep}{.35cm}
	\caption{
		\xt{Comparing the {\em open-world} ReID performance.
			Metric: Detection and Identification Rate (DIR, Eqn. \eqref{eqn:DIR} with $k=1$) 
			over four False Accept Rates (FAR, Eqn. \eqref{eqn:FAR}).}
	}
	\vskip -0.2cm
	\label{tab:opensetRank1}
	\vspace{0.2cm}
	\begin{tabular}{l|cccc|cccc}
		\hline
		Dataset
		& \multicolumn{4}{c|}{PRID$2011$ \cite{hirzer11a}}
		& \multicolumn{4}{c}{iLIDS-VID \cite{wang2014person}}
		\\
		\hline%\noalign{\smallskip}
		$\text{FAR}$ (\%)
		& 1 & 10 & 50 & 100
		& 1 & 10 & 50 & 100
		%& 1 & 10 & 50 & 100
		\\
		\hline \hline
		$L_1$-norm	
		&4.3&8.7&18.5&28.3
		&1.0&5.2&15.6&22.9
		%&6.3&6.3&21.7&31.3
		\\
		%\hline
		MS-SDALF \cite{farenzena2010person}	
		&0.5&1.0&4.5&6.3
		&0.2&0.5&3.3&8.4
		%&3.8&3.8&10.4&15.4
		\\
		%\hline
		\xt{ISR} \cite{lisanti2014person}
		&0.0&18.0&18.2&18.8
		%&8.9&8.9&8.9&10.6
		&0.0&8.9&8.9&10.6
		%&0.0&10.4&14.6&15.8
		\\
		%\hline
		eSDC \cite{zhao2013unsupervised}
		&{5.2}&9.7&20.8&28.3
		&1.4&4.2&8.3&12.4
		%&0.0&20.6&22.6&25.9
		\\
		% \hline
		RDL \cite{ElyorBMVC15}
		& 9.3 & 13.3 & 27.5 & 33.0
		& 2.1 & 4.9 & 10.4 & 13.9
		%& 0.0 &19.4&20.9&23.8
		\\
		%\hline
		MS-ColLBP \cite{hirzer2012relaxed}
		&4.3&6.7&24.3&{39.8}
		&1.1&4.8&15.6&25.9
		%&24.7&24.7&26.2&29.7
		\\
		%\hline
		DVR \cite{WangDVRpami}
		& 4.0 & {12.3} & {34.7} & {46.8}
		& {4.2} & {\bf 14.1} & {\bf 31.8} & {\bf 43.7}
		%&2.9&5.0&11.2&17.9
		\\
		%\hline
		\xt{KCVDCA} \cite{chen2015CVDCA}
		& 14.5 & 20.2 & {\bf 43.0} &49.5
		&{\bf 7.1}&12.1&20.8&24.8
		%&-&-&-&-
		\\
		{XQDA} \cite{liao2015person}
		& 11.5 & 19.8 & {40.3} & {\bf 51.7}
		& 1.3 & 4.3 & 11.5 & 21.2
		%&{\bf 32.4}&{\bf 37.1}&{\bf 57.1}&{\bf 58.5}
		\\
		%\hline
		{\bf MDTS-DTW$_\text{I}$}%51012+histeql
		&{\bf 17.5}&{\bf 25.5}&{38.2}&{46.5}
		&3.4&{8.7}&{26.4}&{37.0}
		%&0.0&29.7&30.3&34.1
		\\
		\hline
		\hline
		%==================================================
		eSDC+DVR
		& 13.3 & 25.2 & 43.3 & 48.5
		& 7.2  & 14.4 & 27.7 & 34.6
		%&9.1&15.3&28.8&36.8
		\\
		eSDC+{\bf MDTS-DTW$_\text{I}$} %51012+histeql
		& 16.8 & 28.2 & 44.7 & 51.3
		& 6.3  & 12.0 & 31.5 & 39.6
		%& 17.9&24.1&40.0&47.4
		\\
		%\hline % ****************
		
		\xt{ISR+DVR}
		&15.0&27.8&42.7&47.7
		&10.7&20.3&29.3&32.9
		%&20.0&20.0&35.8&38.7
		\\
		\xt{ISR+{\bf MDTS-DTW$_\text{I}$}}
		&25.2&36.0&46.8&49.7
		&11.3&17.6&32.6&35.7
		%&20.8&20.8&39.6&43.3
		\\
		%\hline % ****************
		RDL+DVR
		& 26.7 & 39.3 & 58.8 & 62.7
		& 8.5  & 15.4 & 30.1 & 37.3
		%&7.4&10.9&27.9&37.1
		\\
		RDL+{\bf MDTS-DTW$_\text{I}$}
		& 21.8 & 38.5 & 59.7 & 63.7
		& 9.2  & 18.7 & 33.7 & 41.7
		%&16.5&21.5&38.8&44.4
		\\
		%\hline % ****************
		MS-ColLBP+DVR
		& 25.5 & 29.2 & 45.8 & 50.0
		& 16.3 & 22.6 & 38.5 & 43.3
		%&11.8&19.4&34.4&40.0
		\\
		MS-ColLBP+{\bf MDTS-DTW$_\text{I}$} %51012+histeql
		& 27.7 & 33.2 & 49.7 & 51.2
		& 11.6 & 21.3 & 43.8 & 50.0
		%&17.1&29.4&46.2&52.4
		\\
		%\hline % ****************
		\xt{KCVDCA+DVR}
		& 31.7 & 55.2 & 72.5 &75.3
		&17.0&29.7&50.9&56.4
		%&-&-&-&-
		\\
		\xt{KCVDCA+{\bf MDTS-DTW$_\text{I}$}}
		& 42.7 & 52.8 & 72.5&73.5
		&16.8&30.2&51.5&56.4
		%&-&-&-&-
		\\
		XQDA+DVR
		& {\bf 46.8} & {\bf 58.3} & {\bf 78.3} & {\bf 79.7}
		& {\bf 17.3} & {29.1} & {49.9} & {\bf 57.8}
		%&13.8&22.9&39.1&45.6
		\\
		XQDA+{\bf MDTS-DTW$_\text{I}$}
		& 42.7 & 55.2 & 70.5 & 72.8
		& 12.7 & {\bf 32.6} & {\bf 51.8} & 57.3
		%&{\bf 32.4}&{\bf 37.1}&{\bf 57.1}&{\bf 58.5}
		\\
		\hline
	\end{tabular}
	%\vspace{-0.4cm}
\end{table*}

%ssssssssssssssssssssssssssssssssssssssssssss
\subsection{Evaluation on Open-World Person ReID}
\label{sec:evaluation-on-open-world-person-reid}

In this section,
we evaluated the open-world ReID problem,
a more practical scenario compared to the above closed-world setting.
Different single ReID methods and their combinations were assessed and reported in Table \ref{tab:opensetRank1}.
\xt{The performance evaluation metric is Detection and Identification Rate (DIR, Eqn. \eqref{eqn:DIR}) with $k=1$ (e.g. Rank-$1$) at given False Accept Rates (FAR, Eqn. \eqref{eqn:FAR}).}
For the performance of single models, largely similar situations are found as in the closed-world case.
\xt{Particularly, for iLIDS-VID, the supervised space-time ReID method DVR obtains the best results followed by our approach and KCVDCA 
but ours is unsupervised.
On PRID$2011$,
our method has the best DIR scores %much better Rank-1 scores than both DVR and XQDA
given low ($\leq 10\%$) FAR rates 
(corresponding to small $\tau$ in Eqn. \eqref{eqn:FAR}).}
That means,
our method can recognise more accurately the true match at rank-1
when the false accept rate is required to be small.
This situation is mostly ignored in the current ReID literature but
very important in real-world applications, particularly when a large number of probe people are given
and high FARs are not acceptable.

When fusing appearance and space-time feature based ReID methods,
the recognition scores across all FARs are greatly improved,
similar to the early observations.
\xt{In particular, the best ReID accuracies are obtained by the combination of XQDA/KCVDCA and DVR/Ours,
assuming truth match labels are accessible.
In the unsupervised setting, RDL+Ours is the best on both PRID$201$ and iLIDS-VID.}
Clearly, most findings in the closed-world scenario
can be reflected in the open-world setting,
whilst some new different observations emerge especially under
strict false accept rate conditions.
In general, all comparisons above extensively validate the advantages
and effectiveness of the proposed video representation and selective matching models
for person ReID.

%----------------------------------------
\section{Conclusion and Future Work}
\label{sec:conclusion}

\noindent {\bf Conclusion}.
In this work, we presented a video matching based person ReID framework.
This is achieved by (1) developing an effective spatio-temporal pyramids based video representation, called Spatio-Temporal Pyramid Sequence (STPS),
for encoding more effective and complete space-time information
available in person video data; and
(2) formulating a novel Time Shift Dynamic Time Warping (TS-DTW) model
and its Multi-Dimensional extension named MDTS-DTW for selective
matching between pairs of inherently
incomplete and noisy image sequences from two disjoint camera views.
Our method also shows significant complementary effect on previous spatial appearance based ReID approaches for obtaining favourable ReID accuracies.
Importantly, our model is unsupervised and does not require
exhaustive cross-view pairwise data annotation for every camera pair
in model building.
Under both the closed-world and open-world ReID scenarios,
extensive comparative evaluations have demonstrated clearly the advantages of the proposed approach over a wide
range of contemporary state-of-the-art gait recognition, temporal sequence
matching, supervised and unsupervised ReID methods.

\vspace{0.1cm}
\noindent {\bf Future work}.
Our future work for the unsolved person ReID problem includes:
(1)
How to introduce other complementary schemes beyond time shift based data selection
for further
suppressing noisy observations caused by background distractions;
(2)
How to exploit effectively extra types of information
(e.g. semantic text from human or correlated sources)
as computing constraints for improving the matching performance.

\section*{Acknowledgements}
%Acknowledgments should be inserted at the end of the paper, before the
%references, not as a footnote to the title. Use the unnumbered
%Acknowledgements Head style for the Acknowledgments heading.
This work was partially supported by
National Basic Research Program of China (973 Project) 2012CB725405,
the national science and technology support program(2014BAG03B01),
National Natural Science Foundation China 61273238,
Beijing Municipal Science and Technology Project (D15110900280000), Tsinghua University Project (20131089307) and the Foundation of Beijing Key Laboratory for Cooperative Vehicle Infrastructure Systems and Safety Control.
Xiatian Zhu and Xiaolong Ma equally contributed to this work.

\bibliography{refs}

\end{document}